# STORE: Sparse Tensor Response Regression and Neuroimaging Analysis

Will Wei Sun[*] and Lexin Li[†]

**Abstract**

Motivated by applications in neuroimaging analysis, we propose a new regression model, Sparse TensOr REsponse regression (STORE), with a tensor response and a vector predictor. STORE embeds two key sparse structures: element-wise sparsity and low-rankness. It can handle both a non-symmetric and a symmetric tensor response, and thus is applicable to both structural and functional neuroimaging data. We formulate the parameter estimation as a non-convex optimization problem, and develop an efficient alternating updating algorithm. We establish a non-asymptotic estimation error bound for the actual estimator obtained from the proposed algorithm. This error bound reveals an interesting interaction between the computational efficiency and the statistical rate of convergence. When the distribution of the error tensor is Gaussian, we further obtain a fast estimation error rate which allows the tensor dimension to grow exponentially with the sample size. We illustrate the efficacy of our model through intensive simulations and an analysis of the Autism spectrum disorder neuroimaging data.

[*]Assistant Professor, Department of Management Science, University of Miami School of Business Administration, Miami, FL 33146. Email: wsun@bus.miami.edu.

[†]Associate Professor, Division of Biostatistics, University of California, Berkeley, Berkeley, CA 94720. Email: lexinli@berkeley.edu. Li's research was partially supported by NSF grant DMS-1613137 and NIH grant AG034570.

# 1 Introduction

In this article, we study a class of high-dimensional regression models with a tensor response and a vector predictor. Tensor, a.k.a. multidimensional array, is now frequently arising in a wide range of scientific and business applications (Yuan and Zhang, 2016). Our motivation comes from neuroimaging analysis. One example is anatomical magnetic resonance imaging (MRI), where the data takes the form of a three-dimensional array, and image voxels correspond to brain spatial locations. Another example is functional magnetic resonance imaging (fMRI), where the goal is to understand brain functional connectivity that is encoded by a symmetric matrix, with rows and columns corresponding to brain regions, and entries corresponding to interaction between those regions. In both examples, it is of keen scientific interest to compare the scans of brains, or the brain connectivity patterns, between the subjects with neurological disorder and the healthy controls, after adjusting for additional covariates such as age and sex. Both can be more generally formulated as a regression problem, with image tensor or connectivity matrix serving as a response, and the group indicator and other covariates forming a predictor vector.

## 1.1 Our proposal

We develop a general class of tensor response regression models, namely STORE, and embed two key sparse structures: element-wise sparsity and low-rankness. Both structures serve to greatly reduce the computational complexity of the estimation procedure. Meanwhile, both are scientifically plausible in plenty of applications, and have been widely employed in high-dimensional multivariate regressions (e.g., Tibshirani, 1996; Zou, 2006; Yuan and Lin, 2006; Peng et al., 2010; Zhou et al., 2013; Raskutti and Yuan, 2016). A unique feature of STORE is that, it can not only handle a non-symmetric tensor response, but also a symmetric tensor response, and thus is applicable to both structural and functional neuroimaging analysis.

We formulate the learning of STORE as a non-convex optimization problem, and accordingly develop an efficient alternating updating algorithm. Our algorithm consists of two major steps, and each step iteratively updates a subset of the unknown parameters while fixing the others. In Step 1, we reformulate the estimation as a sparse tensor decomposition problem and then employ a decomposition algorithm, the truncated tensor power method (Sun et al., 2017), for solution. In Step 2, we utilize the bi-convex structure of the problem to obtain a closed-form solution.

We carry out a non-asymptotic theoretical analysis for the actual estimator obtained from our optimization algorithm. Based upon a set of our newly developed techniques to tackle the non-convexity in estimation, we obtain an explicit error bound of the actual minimizer. Specifically, let $\mathcal{E}_i, i = 1, \ldots, n$, denote the error tensor, and $\boldsymbol{\Theta}^*$ denote the set of all true parameters. Given an initial parameter with an initialization error $\epsilon$, the finite sample error bound of the $t$-th step solution $\widehat{\boldsymbol{\Theta}}^{(t)}$ consists of two parts:

$$D\left(\widehat{\boldsymbol{\Theta}}^{(t)}, \boldsymbol{\Theta}^*\right) \leq \underbrace{\kappa^t \epsilon}_{\text{computational error}} + \underbrace{\frac{1}{1-\kappa} \max \left\{ C \cdot \eta \left(\frac{1}{n} \sum_{i=1}^n \mathcal{E}_i, s\right), \frac{\widetilde{C}}{\sqrt{n}} \right\}}_{\text{statistical error}},$$

with a high probability. Here $\kappa \in (0, 1)$ is a contraction coefficient, $C$ and $\widetilde{C}$ are some positive constants, and $\eta\left(n^{-1} \sum_{i=1}^n \mathcal{E}_i, s\right)$ represents the $s$-sparse spectral norm of the averaged error tensor; see (8) for a formal definition of this norm. This error bound portrays the estimation error in each iteration, and reveals an interesting interplay between the computational efficiency and the statistical rate of convergence. Note that the computational error decays geometrically with the iteration number $t$, whereas the statistical error remains the same when $t$ grows. Therefore, this bound provides a meaningful guidance for the maximal number of iterations $T$. That is, we stop when the computational error becomes dominated by the statistical error. The resulting estimator falls within the statistical precision of the true parameter $\boldsymbol{\Theta}^*$. Additionally, this finite sample error bound provides a general theoretical guarantee of our estimator, and the result holds for any distribution of the error tensor $\mathcal{E}_i$, by noting that it relies on $\mathcal{E}_i$ only through its sparse spectral norm $\eta\left(n^{-1} \sum_{i=1}^n \mathcal{E}_i, s\right)$. In Sections 4.3 and 4.4, we obtain explicit forms of statistical errors when the distribution of $\mathcal{E}_i$ is available. In particular, when the third-order error tensor $\mathcal{E}_i \in \mathbb{R}^{d_1 \times d_2 \times d_3}$, $i = 1, \cdots, n$, follows an i.i.d. Gaussian distribution, we have

$$D\left(\widehat{\boldsymbol{\Theta}}^{(T)}, \boldsymbol{\Theta}^*\right) = O_p\left(\sqrt{\frac{s^3 \log(d_1 d_2 d_3)}{n}}\right),$$

where $s$ is the cardinality parameter of the decomposed components in the tensor coefficients. This fast estimation error rate allows the tensor dimension to grow exponentially with the sample size. When the order of the tensor is one, STORE reduces to the $d$-dimensional vector regression and our statistical error reduces to $O_p(\sqrt{s \log d/n})$, which is known to be minimax optimal (Wang et al., 2014).

## 1.2 Related works and our contributions

Our work is related to but also clearly distinct from a number of recent development in high-dimensional statistical models involving tensor data.

The first is a class of tensor decomposition methods (Chi and Kolda, 2012; Liu et al., 2012; Anandkumar et al., 2014a,b; Yuan and Zhang, 2016; Sun et al., 2017). Tensor decomposition is an *unsupervised* learning method that aims to find the best low-rank approximation of a single tensor. Our proposed tensor response regression, however, is a *supervised* learning method, which aims to estimate the coefficient tensor that characterizes the association between the tensor response and the vector predictor. Although we utilize the sparse tensor decomposition step of Sun et al. (2017) as part of our estimation algorithm, our objective and technical tools involved are completely different. Particularly, because we work with *multiple* tensor samples, the consistency of our estimator is indexed by both the tensor dimension and the sample size. This is different from the sparse tensor decomposition estimator in Sun et al. (2017) that works with a *single* tensor only. Consequently, a new set of large deviation inequalities are needed and derived in our theoretical investigation. Moreover our algorithm is not restricted to any particular decomposition procedure but can be coupled with other decomposition solutions, e.g., Chi and Kolda (2012); Liu et al. (2012).

The second related line of research tackles tensor regression where the response is a scalar and the predictor is a tensor (Zhou et al., 2013; Wang and Zhu, 2016; Yu and Liu, 2016). In those papers, a low-rank structure is often imposed on the coefficient tensor, which is similar to the low-rank principle we also employ. However, they all treated the tensor as a *predictor*, whereas we treat it as a *response*. This difference in modeling approach leads to different focus and interpretation. The tensor predictor regression focuses on understanding the change of a biological outcome as the tensor image varies, while the tensor response regression aims to study the change of the image as the predictors such as the disease status and age vary. When it comes to theoretical analysis, the two models involve utterly different techniques. In a way, their difference is in analogous to that between multi-response regression and multi-predictor regression in the classical vector-valued regression context.

The third and more relevant line of works directly studies regression with a tensor response (Zhu et al., 2009; Li et al., 2011; Rabusseau and Kadri, 2016; Li and Zhang, 2016; Raskutti and Yuan, 2016). Although aiming to address the same type of problem, our proposal is analytically different in several ways. We also numerically compare with some of those alternative solutions in simulations and real data analysis.

In particular, Zhu et al. (2009) considered a $3 \times 3$ symmetric positive definite matrix arising from diffusion tensor imaging as a response, and developed an intrinsic regression model by mapping the Euclidean space of covariates to the Riemannian manifold of positive-definite matrices. Unlike their solution, we consider a non-symmetric or symmetric tensor response in the Euclidean space, and allow the dimension of the tensor response to increase with the sample size. Li et al. (2011) estimated regression parameters by building iteratively increasing neighbors around each voxel and smoothing observations within the neighbors with weights. By contrast, we model all the voxels in an image tensor in a joint fashion.

Rabusseau and Kadri (2016) considered a tensor response model with a low-rank structure. However, no sparsity is enforced in their estimator, and thus their method is not directly applicable for selecting brain subregions that are affected by the disorder. Our approach instead is capable of region selection, and the resulting estimator is much easier to interpret.

Li and Zhang (2016) proposed an envelope-based tensor response model that is notably different from our proposal. First, they utilized a generalized sparsity principle to exploit the redundant information in the tensor response, by seeking linear combinations of the response that are irrelevant to the regression. Our method instead utilizes the element-wise sparsity in terms of the individual entries. As such our method can achieve region selection, whereas their approach cannot. Second, they obtained the $\sqrt{n}$-convergence rate for the *global* minimizer of their objective function when the tensor dimension is fixed. However, their objective function is non-convex and there is no guarantee that the optimization algorithm can find this global minimizer. By contrast, we derive the error rate of the *actual* minimizer of our algorithm at each iteration, and we also permit an increasing tensor dimension. Third, their approach could not directly incorporate the symmetry constraint when the tensor response is symmetric, which is often encountered in functional imaging analysis. To the best of our knowledge, our solution is the first that can simultaneously tackle both a non-symmetric and a symmetric tensor response in a regression setup.

Raskutti and Yuan (2016) developed a class of sparse regression models, under the assumption of Gaussian error, when either or both the response and predictor are tensors. When the error distribution is Gaussian, our error bound matches theirs. However, our bound is obtained for a general error distribution, where the normality is not necessarily required. In addition, they required another crucial condition that the regularizer must be convex and weakly decomposable. We do not impose this assumption, but instead tackle a non-convex optimization problem, and employ a different set of proof techniques. Finally, they achieved the low-rankness of the estimator through a tensor nuclear norm, which is known to be computationally NP-hard (Friedland and Lim, 2014). By contrast, our rate is established for the actual estimator obtained from our optimization algorithm, which we show later is both feasible and scalable.

Our major contribution is two-fold. First, we propose a new class of regression models with tensor as a response. This model is useful for a wide range of scientific applications, but has received only limited attention in the statistics literature, especially for the symmetric tensor response case. Our proposal is shown to exhibit numerous advantages compared to the existing solutions. Second, we develop a set of new tools for the theoretical analysis of non-convex optimization, which is notably different from recent development in this area (Wang et al., 2014; Yi and Caramanis, 2015; Sun et al., 2015; Balakrishnan et al., 2016). A common technique used in the non-convex optimization analysis is to separately establish the convergence for the population and sample optimizers then combine the two. By contrast, our analysis hinges on exploitation of the bi-convex structure of the objective function, as well as a careful characterization of the impact of the intermediate sparse tensor decomposition step on the estimation error in each iteration step. The bi-convex structure frequently arises in many optimization problems. As such the tools we develop are also of independent interest, and enrich and expand the current toolbox of non-convex optimization analysis.

## 1.3 Notations and structure

Throughout the article we adopt the following notations. Denote $[d] = \{1, \ldots, d\}$, and $\mathbf{I}_d$ the $d \times d$ identity matrix. Denote $1(\cdot)$ the indicator function, and $\circ$ the outer product between vectors. For a vector $\mathbf{a} = \mathbb{R}^d$, $\|\mathbf{a}\|$ denotes its Euclidean norm, and $\|\mathbf{a}\|_0$ the $L_0$ norm, i.e., the number of nonzero entries in $\mathbf{a}$. For a matrix $\mathbf{A} \in \mathbb{R}^{d \times d}$, $\|\mathbf{A}\|$ denotes its spectral norm. For a tensor $\mathcal{A} \in \mathbb{R}^{d_1 \times \cdots \times d_m}$, and a set of vectors $\mathbf{a}_j \in \mathbb{R}^{d_j}, j = 1, \ldots, m$, the multilinear combination of the tensor entries is defined as $\mathcal{A} \times_1 \mathbf{a}_1 \times_2 \ldots \times_m \mathbf{a}_m := \sum_{i_1 \in [d_1]} \cdots \sum_{i_m \in [d_m]} a_{i_1} \ldots a_{i_m} \mathcal{A}_{i_1, \ldots, i_m} \in \mathbb{R}$. The tensor spectral norm is defined as $\|\mathcal{A}\| := \sup_{\|\mathbf{a}_1\| = \ldots = \|\mathbf{a}_m\| = 1} |\mathcal{A} \times_1 \mathbf{a}_1 \times_2 \ldots \times_m \mathbf{a}_m|$, and the tensor Frobenius norm as $\|\mathcal{A}\|_F := \sqrt{\sum_{i_1, \ldots, i_m} \mathcal{A}_{i_1, \ldots, i_m}^2}$. For two tensors $\mathcal{A}, \mathcal{B} \in \mathbb{R}^{d_1 \times \cdots \times d_m}$, their inner product is $\langle \mathcal{A}, \mathcal{B} \rangle = \sum_{i_1, \ldots, i_m} \mathcal{A}_{i_1, \ldots, i_m} \mathcal{B}_{i_1, \ldots, i_m}$. See Kolda and Bader (2009) for more tensor operations. We say $a_n = o(b_n)$ if $a_n/b_n$ converges to 0 as $n$ increases to infinity.

The rest of the article is organized as follows. Section 2 introduces the proposed sparse tensor response regression model, and Section 3 presents the optimization algorithm. Section 4 establishes the estimation error bound. Section 5 presents the simulation results, and Section 6 applies our method to an analysis of the Autism spectrum disorder imaging data. The appendix collects all technical proofs.

## 2 Model

For an $m$th-order tensor response $\mathcal{Y}_i \in \mathbb{R}^{d_1 \times \cdots \times d_m}$, and a vector of predictors $\mathbf{x}_i \in \mathbb{R}^p$, $i = 1, \ldots, n$, we consider the tensor response regression model of the form,

$$\mathcal{Y}_i = \mathcal{B}^* \times_{m+1} \mathbf{x}_i + \mathcal{E}_i, \tag{1}$$

where $\mathcal{B}^* \in \mathbb{R}^{d_1 \times \cdots \times d_m \times p}$ is an $(m+1)$th-order tensor coefficient, and $\mathcal{E}_i \in \mathbb{R}^{d_1 \times \cdots \times d_m}$ is an error tensor independent of $\mathbf{x}_i$. Without loss of generality, the intercept is set to zero, by centering the samples $\mathbf{x}_i$ and $\mathcal{Y}_i$. We also use $d_{m+1} := p$ to represent the dimension of the predictor vector $\mathbf{x}_i$. Our goal is to estimate $\mathcal{B}^*$ given $n$ i.i.d. observations $\{(\mathbf{x}_i, \mathcal{Y}_i), i = 1, \ldots, n\}$.

To facilitate estimation of the ultrahigh dimensional unknown parameters under a limited sample size, it is crucial to introduce some sparse structures. Two most commonly used structures are element-wise sparsity and low-rankness (Raskutti and Yuan, 2016). In the context of tensor response regression, we assume that $\mathcal{B}^*$ admits the following rank-$K$ decomposition structure (Kolda and Bader, 2009),

$$\mathcal{B}^* = \sum_{k \in [K]} w_k^* \boldsymbol{\beta}_{k,1}^* \circ \cdots \circ \boldsymbol{\beta}_{k,m}^* \circ \boldsymbol{\beta}_{k,m+1}^*, \quad w_k^* \in \mathbb{R}, \; \boldsymbol{\beta}_{k,j}^* \in \mathbb{S}^{d_j}, \tag{2}$$

where $\mathbb{S}^d = \{\mathbf{v} \in \mathbb{R}^d \,|\, \|\mathbf{v}\| = 1\}$, $\|\boldsymbol{\beta}_{k,j}^*\|_0 \leq d_{j0} < d_j$, and $d_{j0}$ represents the true sparsity of the individual components $\boldsymbol{\beta}_{k,j}^*$, for $k \in [K], j \in [m+1]$. Consequently, when coupling with

the low-rank structure of (2), the element-wise sparsity of $\boldsymbol{\beta}^*_{k,j}$ implies that some individual entries of $\mathcal{B}^*$ are zero. Moreover, we assume $w^*_{\max} = w^*_1 \geq \cdots \geq w^*_K = w^*_{\min} > 0$, and assume each $w^*_i$ to be bounded away from 0 and $\infty$. Under the structure of (2), estimating $\mathcal{B}^*$ reduces to the estimation of $w^*_k, \beta^*_{k,1}, \ldots, \beta^*_{k,m+1}$ for any $k \in [K]$, and the corresponding parameter space becomes $\mathbb{B} := \{w_k \in \mathbb{R}, \boldsymbol{\beta}_{k,j} \in \mathbb{R}^{d_j}, k \in [K], j \in [m+1] \,|\, c_1 \leq |w_k| \leq c_2, \|\boldsymbol{\beta}_{k,j}\|_2 = 1, \|\boldsymbol{\beta}_{k,j}\|_0 \leq d_{j0}\}$. Accordingly, the number of unknown parameters is reduced from $p \prod_{j=1}^m d_j$ to $K(\sum_{j=1}^m d_j + p)$. To estimate those unknown parameters, we propose to solve the following constrained optimization problem,

$$\min_{\substack{w_k, \boldsymbol{\beta}_{k,1},\ldots,\boldsymbol{\beta}_{k,m+1} \\ k \in [K], j \in [m+1]}} \frac{1}{n} \sum_{i=1}^n \left\| \mathcal{Y}_i - \sum_{k \in [K]} w_k(\boldsymbol{\beta}^\top_{k,m+1}\mathbf{x}_i)\boldsymbol{\beta}_{k,1} \circ \cdots \circ \boldsymbol{\beta}_{k,m} \right\|^2_F, \quad (3)$$

$$\text{subject to } \|\boldsymbol{\beta}_{k,j}\|_2 = 1, \|\boldsymbol{\beta}_{k,j}\|_0 \leq s_j,$$

where $s_j$ is the cardinality parameter of the $j$-th component. Here we encourage the sparsity of the decomposed components via a hard-thresholding penalty to control $\|\boldsymbol{\beta}_{k,j}\|_0$. Compared to the lasso type penalized approach in sparse learning, the hard-thresholding method avoids bias and has been shown to be more appealing in numerous high-dimensional learning problems (Shen et al., 2012, 2013; Wang et al., 2014).

## 3 Estimation

The problem in (3) is a non-convex optimization. The key of our estimation procedure is to explore its *bi-convex* structure. In particular, the objective function in (3) is bi-convex in $(\boldsymbol{\beta}_{k,1}, \ldots, \boldsymbol{\beta}_{k,m+1})$, $k \in [K]$, in that it is convex in $\boldsymbol{\beta}_{k,j}$ when all other parameters are fixed. Utilizing this property, we propose an efficient alternating updating algorithm to solve (3).

### 3.1 Algorithm

We first summarize our estimation procedure in Algorithm 1, then present the two key steps.

**Step 1**: The first core step of our algorithm is to update $w_k, \boldsymbol{\beta}_{k,1}, \ldots, \boldsymbol{\beta}_{k,m}$ for each $k = 1, \ldots, K$, given $\boldsymbol{\beta}_{j,m+1}$, $j = 1, \ldots, K$ and $w_{k'}, \beta_{k',1}, \ldots, \beta_{k',m}$, $k' \neq k$. Letting $\alpha_{ik} := \boldsymbol{\beta}^\top_{k,m+1}\mathbf{x}_i$, $i = 1, \ldots, n$, we note that (3) is equivalent to

$$\min_{\substack{w_k, \boldsymbol{\beta}_{k,1},\ldots,\boldsymbol{\beta}_{k,m} \\ \|\boldsymbol{\beta}_{k,j}\|_2 = 1, \|\boldsymbol{\beta}_{k,j}\|_0 \leq s_j, j \in [m]}} \frac{1}{n} \sum_{i=1}^n \alpha^2_{ik} \left\| \mathcal{R}_i - w_k \boldsymbol{\beta}_{k,1} \circ \cdots \circ \boldsymbol{\beta}_{k,m} \right\|^2_F, \quad (4)$$

where the residual tensor term $\mathcal{R}_i$ is of the form,

$$\mathcal{R}_i := \left[ \mathcal{Y}_i - \sum_{k' \neq k, k' \in [K]} w_{k'}\alpha_{ik'}\boldsymbol{\beta}_{k',1} \circ \cdots \circ \boldsymbol{\beta}_{k',m} \right] / \alpha_{ik}. \quad (5)$$

**Algorithm 1** Alternating updating algorithm for STORE.
1: **Input:** data $\{(\mathbf{x}_i, \mathcal{Y}_i), i = 1, \ldots, n\}$, rank $K$, cardinality vector $(s_1, \ldots, s_m)$.
2: **Initialize** $w_k = 1$ and random unit-norm vectors $\boldsymbol{\beta}_{k,1}, \ldots, \boldsymbol{\beta}_{k,m+1}$ for each $k \in [K]$.
3: **Until** the termination condition is met, **Do**
4:   Step 1: For $k = 1$ to $K$, obtain $\widehat{w}_k, \widehat{\boldsymbol{\beta}}_{k,1}, \ldots, \widehat{\boldsymbol{\beta}}_{k,m}$ by solving (6) via the sparse tensor decomposition procedure of Sun et al. (2017) with parameters $K$ and $(s_1, \ldots, s_m)$.
5:   Step 2: For $k = 1$ to $K$, obtain $\widehat{\boldsymbol{\beta}}_{k,m+1}$ using Lemma 2.
6: **Output:** $\widehat{w}_k, \widehat{\boldsymbol{\beta}}_{k,1}, \ldots, \widehat{\boldsymbol{\beta}}_{k,m+1}$ for each $k \in [K]$.

The next lemma shows that (4) can be solved via an efficient sparse tensor decomposition procedure.

**Lemma 1.** *Denote* $(\widehat{w}_k, \widehat{\boldsymbol{\beta}}_{k,1}, \ldots, \widehat{\boldsymbol{\beta}}_{k,m})$ *as the solution of* (4), *and denote* $(\widetilde{w}_k, \widetilde{\boldsymbol{\beta}}_{k,1}, \ldots, \widetilde{\boldsymbol{\beta}}_{k,m})$ *as the solution of*

$$\min_{\substack{w_k, \boldsymbol{\beta}_{k,1}, \ldots, \boldsymbol{\beta}_{k,m} \\ \|\boldsymbol{\beta}_{k,j}\|_2 = 1, \|\boldsymbol{\beta}_{k,j}\|_0 \le s_j, j \in [m]}} \left\| \bar{\mathcal{R}} - w_k \boldsymbol{\beta}_{k,1} \circ \cdots \circ \boldsymbol{\beta}_{k,m} \right\|_F^2, \tag{6}$$

*where* $\bar{\mathcal{R}} := \frac{1}{n} \sum_{i=1}^n \alpha_{ik}^2 \mathcal{R}_i$. *Then it satisfies that* $\widehat{w}_k = n\widetilde{w}_k / \sum_{i=1}^n \alpha_{ik}^2$ *and* $\widehat{\boldsymbol{\beta}}_{k,l} = \widetilde{\boldsymbol{\beta}}_{k,l}$, *for* $l = 1, \ldots, m$ *and* $k = 1, \ldots, K$.

Lemma 1 implies that the optimization problem in (4) reduces to a sparse rank-one tensor decomposition on the averaged tensor $\bar{\mathcal{R}}$. To efficiently solve (6), in this paper we employ a truncation-based sparse tensor decomposition procedure (Sun et al., 2017), by first solving the non-sparse tensor decomposition components and then truncating them to achieve the desirable sparsity. It is also noteworthy that our method is flexible in the choice of optimization algorithm for solving (6), and Algorithm 1 can be coupled with many other sparse tensor decomposition algorithms, e.g., Chi and Kolda (2012); Liu et al. (2012).

When the tensor response $\mathcal{Y}_i$ is symmetric, the resulting coefficient $\mathcal{B}$ should be symmetric too. Our algorithm can easily adapt to this scenario, by setting $\boldsymbol{\beta}_{k,1} = \cdots = \boldsymbol{\beta}_{k,m} = \boldsymbol{\beta}_k$ for each $k \in [K]$, and the cardinality parameters $s_1 = \cdots = s_m = s$. That is, we slightly modify Step 1 in Algorithm 1 as: for $k = 1$ to $K$, compute $\mathcal{R}_i$ in (5) and solve $\widehat{w}_k, \widehat{\boldsymbol{\beta}}_k$ via the symmetric sparse tensor decomposition with parameters $K$ and $s$.

**Step 2**: The second core step of our algorithm is to update $\boldsymbol{\beta}_{k,m+1}$ for each $k = 1, \ldots, K$, given $w_j, \boldsymbol{\beta}_{j,1}, \ldots, \boldsymbol{\beta}_{j,m}, j = 1, \ldots, K$ and $\boldsymbol{\beta}_{k',m+1}, k' \ne k$. Letting $\mathcal{A}_k = w_k \boldsymbol{\beta}_{k,1} \circ \cdots \circ \boldsymbol{\beta}_{k,m}$, then (3) is equivalent to

$$\min_{\alpha} \frac{1}{n} \sum_{i=1}^n \left\| \mathcal{T}_i - \alpha^\top \mathbf{x}_i \mathcal{A}_k \right\|_F^2, \tag{7}$$

where the residual tensor $\mathcal{T}_i = \mathcal{Y}_i - \sum_{k' \ne k, k' \in [K]} w_{k'} (\boldsymbol{\beta}_{k',m+1}^\top \mathbf{x}_i) \boldsymbol{\beta}_{k',1} \circ \cdots \circ \boldsymbol{\beta}_{k',m}$. The next lemma gives a closed-form solution of (7).

**Lemma 2.** *The solution of* (7) *is given by*

$$\widehat{\boldsymbol{\beta}}_{k,m+1} = \left(\frac{1}{n}\sum_{i=1}^{n}\mathbf{x}_i\mathbf{x}_i^\top\right)^{-1} \frac{n^{-1}\sum_{i=1}^{n}\langle\mathcal{T}_i,\mathcal{A}_k\rangle\mathbf{x}_i}{\|\mathcal{A}_k\|_F^2}.$$

In our neuroimaging example in Section 6, the dimension of the predictor vector is $p = 3$. For such a small value of $p$, the sample size $n$ is generally much larger, and we can directly invert the sample covariance matrix $n^{-1}\sum_{i=1}^{n}\mathbf{x}_i\mathbf{x}_i^\top$. If $n < p$, this matrix is not invertible. Then one may employ sparse graphical model estimation (Yuan and Lin, 2007; Friedman et al., 2008; Zhang and Zou, 2014) and replace this inverse with a sparse precision matrix estimator. Alternatively, one may also introduce additional sparsity constraint on the parameter $\boldsymbol{\beta}_{k,m+1}$ and resort to a regularized estimation approach to update $\boldsymbol{\beta}_{k,m+1}$.

Finally, we terminate the alternating update of Steps 1 and 2 when the new estimates are close to the ones from the previous iteration. The termination condition is set as

$$\max_{j\in[m+1],k\in[K]}\min\left\{\|\widehat{\boldsymbol{\beta}}_{k,j}^{(t)} - \widehat{\boldsymbol{\beta}}_{k,j}^{(t-1)}\|, \|\widehat{\boldsymbol{\beta}}_{k,j}^{(t)} + \widehat{\boldsymbol{\beta}}_{k,j}^{(t-1)}\|\right\} \le 10^{-4}.$$

## 3.2 Tuning parameter selection

In Algorithm 1, the rank $K$ and the cardinality $s_1, \ldots, s_m$ are tuning parameters. We propose to select those parameters via a BIC-type criterion. Specifically, given a pre-specified set of rank values $\mathcal{K}$ and a pre-specified set of cardinality values $\mathcal{S}_1, \ldots, \mathcal{S}_m$, we choose the combination of parameters $(\widehat{K}, \widehat{s}_1, \ldots, \widehat{s}_m)$ that minimizes

$$\mathrm{BIC} = \log\left(\sum_{i=1}^{n}\|\mathcal{Y}_i - \widehat{\mathcal{B}}\times_{m+1}\mathbf{x}_i\|_F^2\right) + \frac{\log(n\prod_{j=1}^{m}d_m)}{n\prod_{j=1}^{m}d_m}\sum_{k=1}^{K}\sum_{j=1}^{m}\|\widehat{\boldsymbol{\beta}}_{k,j}\|_0.$$

This criterion balances between model fitting and model sparsity, and a similar version has been commonly employed in rank estimation (Zhou et al., 2013).

## 4 Theory

Next we establish the error bound of the actual STORE estimator obtained from our Algorithm 1. The resulting error bound consists of two quantities: a computational error and a statistical error. The computational error captures the error caused by the non-convexity of the optimization problem, whereas the statistical error measures the error due to finite samples.

In order to compute the distance between the estimator and the truth, we define the distance measure between two unit vectors $\mathbf{u}, \mathbf{v} \in \mathbb{R}^d$ as $D(\mathbf{u}, \mathbf{v}) := \sqrt{1-(\mathbf{u}^\top\mathbf{v})^2}$. We then have $D(\mathbf{u}, \mathbf{v}) \le \min\{\|\mathbf{u}-\mathbf{v}\|, \|\mathbf{u}+\mathbf{v}\|\} \le \sqrt{2}D(\mathbf{u}, \mathbf{v})$. The distance function $D(\mathbf{u}, \mathbf{v})$ resolves

the sign issue in the decomposition components since changing the signs of any two component vectors while fixing other component vectors does not affect the generated tensor.

## 4.1 Assumptions

We first introduce the technical assumptions required to guarantee the desirable error bound. The first assumption is about the structure of the true model.

**Assumption 1** (Model Assumption). *In model* (1), *we assume the true coefficient $\mathcal{B}^*$ is sparse and low-rank satisfying* (2) *and such decomposition is unique up to a permutation. Assume $\|\mathcal{B}^*\| \leq C_1 w_{\max}^*$, and for each $i$, assume $\|\mathbf{x}_i\| \leq C_2$ and $|\boldsymbol{\beta}_{k,m+1}^{*\top} \mathbf{x}_i| \geq C_3$ almost surely, for some positive constants $C_1, C_2, C_3$. Furthermore, we require the eigenvalues of the sample covariance $\Sigma := n^{-1} \sum_{i=1}^{n} \mathbf{x}_i \mathbf{x}_i^\top$ to satisfy $c_0 < \lambda_{\min}(\Sigma) \leq \lambda_{\max}(\Sigma) < \widetilde{c}_0$ for some constants $c_0, \widetilde{c}_0$.*

The above unique decomposition condition is to ensure the identifiability of tensor decomposition. Kruskal (1976, 1977) provided the classical conditions for such identifiability if the sum of the Kruskal ranks of the $m$ decomposed component matrices is larger than $2K+2$. In our model, the rank $K$ is fixed, and hence the identifiability of our tensor decomposition (2) is guaranteed when the decomposed components are not highly dependent. Moreover, the conditions on the predictor vector $\mathbf{x}_i$ are mild and trivially hold when the dimension of $\mathbf{x}_i$ is fixed. For instance, in Section 6, the dimension of $\mathbf{x}_i$ is 3 in our neuroimaging data example.

The second assumption is about the initialization in Algorithm 1.

**Assumption 2** (Initialization Assumption). *Define the initialization error*

$$\epsilon := \max \left\{ \max_k \|\widehat{w}_k^{(0)} - w_k^*\|_2, \max_{k,j} D\left(\widehat{\boldsymbol{\beta}}_{k,j}^{(0)}, \boldsymbol{\beta}_{k,j}^*\right) \right\}.$$

*We assume that*

$$\epsilon < \min \left\{ \sqrt{\frac{w_{\min}^*}{2(w_{\min}^* + w_{\max}^* C_1)}}, \frac{w_{\min}^{*2}}{8\sqrt{5} w_{\max}^* C_1 (6\sqrt{2} w_{\min}^* + 2)}, \frac{C_3}{2C_2}, \frac{w_{\min}^*}{2} \right\},$$

*where the constants $C_1, C_2, C_3$ are as defined in Assumption 1.*

Note that our assumption on the initialization parameters only requires the error to be bounded by some constant; i.e., the initial estimators are not too far away from the true parameters. This assumption is necessary to handle the non-convexity of the optimization, and has been commonly imposed in non-convex optimization (Wang et al., 2014; Balakrishnan et al., 2016). It is also noteworthy that this assumption is satisfied by the estimators from sparse singular value decomposition of the unfolding matrix from the original tensor in Step 1 of Algorithm 1; see Sun et al. (2017) for more discussion on the theoretical guarantee of such an initialization procedure.

The third assumption requires that the error tensor $\mathcal{E}_i$ is controlled. Before stating the assumption, we introduce a critical concept to measure the noise level in the tensor response regression. In particular, we define the sparse spectral norm of $\mathcal{E}$ as

$$\eta(\mathcal{E}, d_{01}, \cdots, d_{0m}) := \sup_{\substack{\|\mathbf{u}_1\|=\cdots=\|\mathbf{u}_m\|=1 \\ \|\mathbf{u}_1\|_0 \leq d_{01}, \ldots, \|\mathbf{u}_m\|_0 \leq d_{0m}}} \left| \mathcal{E} \times_1 \mathbf{u}_1 \times_2 \cdots \times_m \mathbf{u}_m \right|. \quad (8)$$

This quantifies the perturbation error in a sparse scenario, and in the sparse case with $d_{0j} \ll d_j$ ($j = 1, \ldots, m$), it is much smaller than the spectral norm $\|\mathcal{E}\|$, i.e., $\eta(\mathcal{E}, d_1, \cdots, d_m)$. Denoting $d_0 = \max_j d_{0j}$, we have $\eta(\mathcal{E}, d_{01}, \cdots, d_{0m}) \leq \eta(\mathcal{E}, d_0, \cdots, d_0)$. We write $\eta(\mathcal{E}, d_0) := \eta(\mathcal{E}, d_0, \cdots, d_0)$.

**Assumption 3** (Error Assumption). *Let $s := \max\{s_1, \ldots, s_m\}$, with $s_j$ being the cardinality parameter in Algorithm 1. We assume the error tensor $\mathcal{E}_i$, $i = 1, \ldots, n$, satisfies that*

$$\eta\left(\frac{1}{n} \sum_{i=1}^n \mathcal{E}_i, s\right) \leq \frac{C_3 w^*_{\min}}{8},$$

*where $C_3$ is as defined in Assumption 1.*

This assumption requires that the observed tensor responses, $\mathcal{Y}_i$, $i = 1, \ldots, n$, are not too noisy. It is a minor condition, since we only require the spectral norm of the error tensor to be bounded. As we later show in Section 4.3, when $\mathcal{E}_i$ is a standard Gaussian tensor, its spectral norm converges to zero when the sample size $n$ increases.

## 4.2 Main result

Next we present our main theorem that describes the estimation error of the actual estimator at each iteration from Algorithm 1.

**Theorem 1.** *Assuming Assumptions 1, 2, and 3, the estimator in the t-th iteration of Algorithm 1, with $s \geq d_0 := \max_j d_{j0}$, satisfies that*

$$\max\left\{\max_k \|\widehat{w}_k^{(t)} - w_k^*\|_2, \max_{k,j} D\left(\widehat{\boldsymbol{\beta}}_{k,j}^{(t)}, \boldsymbol{\beta}_{k,j}^*\right)\right\}$$

$$\leq \underbrace{\kappa^t \epsilon}_{computational\ error} + \underbrace{\frac{1}{1-\kappa} \max\left\{\frac{4\sqrt{5}}{C_3 w^*_{\min}} \eta\left(\frac{1}{n}\sum_{i=1}^n \mathcal{E}_i, s\right), \frac{\widetilde{C}}{\sqrt{n}}\right\}}_{statistical\ error}, \quad (9)$$

*where the contraction coefficient $\kappa = (\kappa_1 + \kappa_2)\kappa_3 \in (0,1)$, with $\kappa_1 := 4\sqrt{5} w^*_{\max} C_3 \epsilon / w^*_{\min}$, $\kappa_2 := 2C_2 \eta\left(\bar{\mathcal{E}}, s\right)/C_3^2$, and $\kappa_3 := 2/w^*_{\min} + 6\sqrt{2}$. Here $C_1, C_2, C_3$ are constants as defined in Assumption 1, and $\widetilde{C}$ is some positive constant.*

Theorem 1 reveals the interplay between the computational error and the statistical error for our estimator, and has some interesting implications. First, it provides a theoretical guidance to terminate the iterations of the alternating updating algorithm. That is, when the computational error is dominated by the statistical error, in that

$$t \geq T := \log_{1/\kappa} \left( \frac{(1-\kappa)\epsilon}{\max\left\{ \frac{4\sqrt{5}}{C_3 w^*_{\min}} \eta\left(\frac{1}{n}\sum_{i=1}^n \mathcal{E}_i, s\right), \frac{\widetilde{C}}{\sqrt{n}} \right\}} \right), \tag{10}$$

the estimator from our algorithm achieves an error at the rate

$$O_p\left( \max\left\{ \eta\left(\frac{1}{n}\sum_{i=1}^n \mathcal{E}_i, s\right), \frac{1}{\sqrt{n}} \right\} \right). \tag{11}$$

The constants $C_3$ and $\widetilde{C}$ in (10) are unknown in practice. However, due to the fact both constants are independent of the sample size, the sparsity parameters, and the dimensionality, the expression (10) gives an order of the optimal termination $T$ in terms of the parameters that go to infinity. Second, Theorem 1 also provides a general theoretical guarantee for our estimator for any distribution of the error tensor $\mathcal{E}_i$, as the error rate depends on the distribution of the error tensor only through $\eta\left(\frac{1}{n}\sum_{i=1}^n \mathcal{E}_i, s\right)$. Later we obtain the explicit forms of the error rates for some special error distributions.

We also remark that, Theorem 1 is obtained assuming the true rank $K$ is known. This is a common assumption in theoretical analysis in the tensor literature (Zhou et al., 2013; Li and Zhang, 2016). To our knowledge, none of the exiting works has provided a provable estimation of the rank, due to that the exact tensor rank calculation is NP hard (Kolda and Bader, 2009). The theory of rank estimation will be our future research.

### 4.3 Gaussian tensor error

We first derive the explicit form of the statistical error in Theorem 1 when $\mathcal{E}_i, i = 1, \ldots, n$, are i.i.d. Gaussian tensors; i.e., the entries of $\mathcal{E}_i$ are i.i.d. standard Gaussian random variables.

**Corollary 1.** *Under the assumptions of Theorem 1, and assuming $\mathcal{E}_i \in \mathbb{R}^{d_1 \times d_2 \times d_3}, i = 1, \ldots, n$, are i.i.d. Gaussian tensors, we have*

$$\eta\left(\frac{1}{n}\sum_{i=1}^n \mathcal{E}_i, s\right) = O_p\left( \sqrt{\frac{s^3 \log(d_1 d_2 d_3)}{n}} \right).$$

*Consequently the estimator at the $T$-th iteration of Algorithm 1 satisfies*

$$\max\left\{ \max_k \|\widehat{w}_k^{(T)} - w_k^*\|_2, \max_{k,j} D\left(\widehat{\boldsymbol{\beta}}_{k,j}^{(T)}, \boldsymbol{\beta}_{k,j}^*\right) \right\} = O_p\left( \sqrt{\frac{s^3 \log(d_1 d_2 d_3)}{n}} \right).$$

We provide some insight about the statistical error $O_p(\sqrt{s^3 \log(d_1 d_2 d_3)/n})$. Note that, $s$ is the maximal cardinality of vectors $\widehat{\boldsymbol{\beta}}_{k,j}$ for $k \in [K]$ and $j \in [m]$, and hence there are at most $O(s^3)$ non-zero elements and $O(d_1 d_2 d_3)$ free parameters in the true tensor coefficient $\mathcal{B}^*$. This error rate allows the tensor dimension to grow exponentially with the sample size and matches the one established by Raskutti and Yuan (2016) for sparse tensor regression. When the order of tensor is one, i.e., $m = 1$, it reduces to the statistical error $O_p(\sqrt{s \log d/n})$ for the high-dimensional vector regression $\mathbf{y} = \mathbf{X}\boldsymbol{\beta} + \epsilon$ with $\boldsymbol{\beta} \in \mathbb{R}^d$ and $\|\boldsymbol{\beta}\|_0 \leq s$, and is known to be minimax optimal (Wang et al., 2014).

## 4.4 Symmetric matrix error

Next we consider the case when the response $\mathcal{Y}_i$ is a symmetric matrix ($m = 2$). Such a scenario is often encountered in functional neuroimaging analysis, where the target of interest is the brain connectivity pattern encoded in the form of a symmetric correlation matrix. The symmetry of the coefficient $\mathcal{B}^*$ requires that $\boldsymbol{\beta}^*_{k,1} = \boldsymbol{\beta}^*_{k,2} = \boldsymbol{\beta}^*_k$. Henceforth, $\mathcal{B}^* = \sum_{k \in [K]} w_k^* \cdot \boldsymbol{\beta}^*_k \circ \boldsymbol{\beta}^*_k \circ \boldsymbol{\beta}^*_{k,3}$, where $w_k^* \in \mathbb{R}$, $\|\boldsymbol{\beta}^*_k\|_2 = 1$, $\|\boldsymbol{\beta}^*_k\|_0 \leq d_0$, and $\boldsymbol{\beta}^*_{k,3} \in \mathbb{S}^p$. To facilitate the derivation of the explicit form of $\eta\left(\frac{1}{n}\sum_{i=1}^n \mathcal{E}_i, s\right)$, we assume that the error matrix $\mathcal{E}_i$ satisfies

$$\mathcal{E}_i = (\widetilde{\mathcal{E}}_i + \widetilde{\mathcal{E}}_i^\top)/2, \tag{12}$$

where $\widetilde{\mathcal{E}}_i \in \mathbb{R}^{d \times d}$ is a matrix whose entries are i.i.d. standard Gaussian. This assumption is mainly for technical reasons, and the theoretical analysis of a more general symmetric tensor response is left as future work.

**Corollary 2.** *Under the assumptions of Theorem 1, and assuming $\mathcal{E}_i, i = 1, \ldots, n$ are i.i.d. and of the form (12), we have*

$$\eta\left(\frac{1}{n}\sum_{i=1}^n \mathcal{E}_i, s\right) = O_p\left(\sqrt{\frac{s^2 \log(d^2)}{n}}\right).$$

# 5 Simulations

In this section, we investigate the numerical performance of our method, and demonstrate its superior performance when compared to some alternative solutions.

To evaluate the estimation accuracy, we report the estimation error $\|\widehat{\mathcal{B}} - \mathcal{B}\|_F$. To evaluate the variable selection accuracy, we compute the true positive rate, $\text{TPR} := m^{-1}\sum_{j=1}^m \text{TPR}_j$, and the false positive rate, $\text{FPR} := m^{-1}\sum_{j=1}^m \text{FPR}_j$, where $\text{TPR}_j := K^{-1}\sum_{k=1}^K \sum_i \mathbf{1}([\beta_{k,j}]_i \neq 0, [\widehat{\beta}_{k,j}]_i \neq 0)/\sum_i \mathbf{1}([\beta_{k,j}]_i \neq 0)$, $\text{FPR}_j := K^{-1}\sum_{k=1}^K \sum_i \mathbf{1}([\beta_{k,j}]_i = 0, [\widehat{\beta}_{k,j}]_i \neq 0)/\sum_i \mathbf{1}([\beta_{k,j}]_i = 0)$. We also report the F1 score, $F_1 := 2(\text{recall}^{-1} + \text{precision}^{-1})^{-1}$, where precision $= m^{-1}K^{-1}\sum_{j=1}^m \sum_{k=1}^K \sum_i \mathbf{1}([\beta_{k,j}]_i \neq 0, [\widehat{\beta}_{k,j}]_i \neq 0)/\sum_i \mathbf{1}([\widehat{\beta}_{k,j}]_i = 0)$ and

recall = TPR. The TPR, FPR and F1 score are all between 0 and 1, with a larger value of TPR and F1 and a smaller value of FPR indicating a better selection performance.

We compare our STORE method with four alternative solutions. The first is ordinary least squares (OLS) that fits one response variable at a time. Although a naive solution, OLS is widely used in neuroimaging analysis due to its simplicity. The second is the sparse ordinary least squares method (Sparse OLS) of Peng et al. (2010) that first vectorizes the tensor response then fits a regularized multivariate regression with the lasso penalty. The third is the envelope-based tensor response regression method (ENV) of Li and Zhang (2016) that utilizes a generalized sparsity principle. It is noteworthy that, in Li and Zhang (2016), ENV has been shown to clearly outperform the tensor predictor regression method of Zhou et al. (2013). Hence, we directly compare with ENV but no longer with Zhou et al. (2013). The fourth is the higher-order low-rank regression method (HOLRR) of Rabusseau and Kadri (2016) that imposes a low-rank tensor structure. We also comment that, OLS and sparse OLS both ignore the tensor structure of the data, whereas neither of OLS, ENV, and HOLRR performs region selection.

## 5.1 3D tensor response example

We first consider an example of a third-order tensor response ($m = 3$). The data was generated from model (1) with $x_i$ a scalar taking values 0 or 1 with an equal probability, and the coefficient tensor $\mathcal{B} \in \mathbb{R}^{d_1 \times d_2 \times d_3}$ of the form, $\mathcal{B} = \sum_{k \in [K]} w_k \beta_{k,1} \circ \beta_{k,2} \circ \beta_{k,3} \circ \beta_{k,4}, \quad w_k \in \mathbb{R}, \beta_{k,j} \in \mathbb{S}^{d_j}$. For each $k \in [K]$, we first generated i.i.d. standard Gaussian entries of the three vectors $\beta_{k,1}, \beta_{k,2}, \beta_{k,3}$, then truncated each vector with the cardinality parameters $s_{01}, s_{02}, s_{03}$ accordingly. Next we normalized each vector and aggregated the coefficients as $w_k$. We set $\beta_{k,4}$ as one. In all simulations, we fixed the tensor dimensions $d_1 = 100, d_2 = 50, d_3 = 20$. We set the true cardinality $s_{0j} = s * d_j$ ($j = 1, 2, 3$), and varied the sparsity level $s \in \{0.3, 0.5\}$ and rank $K \in \{2, 5\}$. We considered the sample size $n \in \{20, 100\}$. Based on our theoretical finding, a larger sample size leads to a better estimation performance. We have chosen a relatively small sample size to mimic the situation where the number of subjects is limited. Such a situation is commonly encountered in neuroimaging studies. In total, there are 8 combinations of different scenarios. For STORE, the tuning parameters were chosen according to the BIC criterion in Section 3.2 by setting $\mathcal{K} = \{1, 2, \ldots, 10\}$ and $\mathcal{S} = \{0.1, 0.2, \ldots, 0.9\}$. The tunings for Sparse OLS and HOLRR were conducted based on their recommended approaches.

Table 1 reports simulation results based on 20 replications. It is seen that our STORE method clearly achieves a superior performance than all the competing methods in terms of both estimation accuracy and variable selection accuracy. The same pattern holds for different sample size $n$, rank $K$, and sparsity level $s$. It is also noted that, since OLS, ENV, and HOLRR do not incorporate entry-wise sparsity, their corresponding TPR and FPR values are always one, and their F1 scores are undefined due to zero in the denominator of the precision.

Figure 1: The computational time of our algorithm as the sample size and the tensor dimension varies. The theoretical linear rate is shown in red triangle.

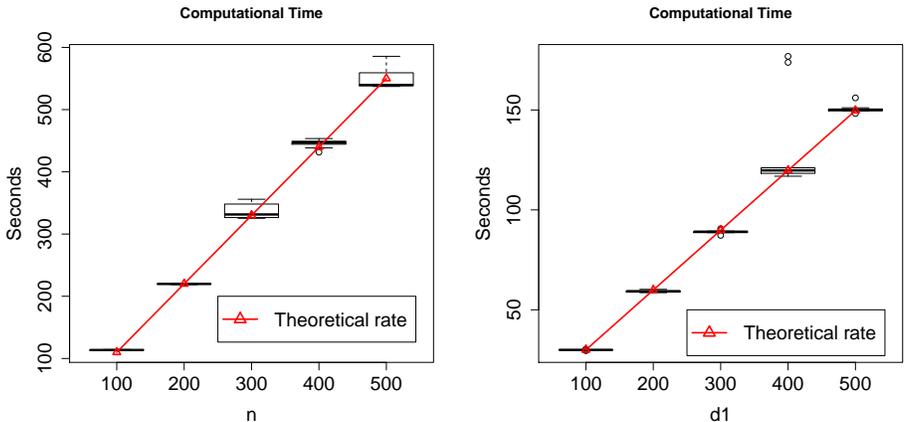

We also studied the computational time of our algorithm when the sample size and the tensor dimension increase. The code is written in R and is implemented on a laptop with 2.5 GHz Intel Core i7 processor. Figure 1 reports the results based on 20 replications. Specifically, we fixed the rank $K = 2$ and the sparsity $s = 0.3$. In the left panel, we fixed the tensor dimensions $d_1 = 100, d_2 = 50, d_3 = 20$, but varied the sample size $n \in \{100, 200, 300, 400, 500\}$. In the right panel, we fixed $n = 20$, $d_2 = 50$, $d_3 = 20$, but varied the dimension $d_1 \in \{100, 200, 300, 400, 500\}$. This scale of sample size and tensor dimension is typical in neuroimaging applications. From the plot we see that the computational time of our Algorithm 1 is roughly linear in terms of the sample size and the tensor dimension when other parameters are fixed. This shows that our algorithm is scalable and feasible with moderate to large sample size and dimensionality. Meanwhile, it is possible to further improve our current algorithm. For instance, in each iteration, one could employ the recently developed parallel tensor decomposition (Papalexakis et al., 2012), or the sketching tensor decomposition (Wang et al., 2015), to accelerate the computation.

## 5.2 2D symmetric matrix response example

Next we considered an example of a second-order symmetric tensor (matrix, $m = 2$). This type of problem is commonly encountered in functional neuroimaging and brain connectivity analysis. The data was generated from the model, $\mathcal{Y}_i = \mathcal{B} \times_{m+1} x_i + \mathcal{E}_i$, where $\mathcal{Y}_i \in \mathbb{R}^{100 \times 100}$ is the symmetric matrix response, $x_i$ is a scalar taking values 0 or 1, and the error term $\mathcal{E}_i$ is a symmetric matrix whose upper triangle entries are generated randomly from a standard normal distribution. The sample size is $n = 20$. To mimic some common connectivity patterns, we have simulated three symmetric coefficient matrices $\mathcal{B}$ that lead to three graph patterns: a randomly generated sparse and low-rank graph (random), a hub graph (hub), and a small-world graph (small-world). Note that the coefficient tensor $\mathcal{B}$ is symmetric, though the corresponding graph pattern is not necessarily so. The random and hub graphs are both of a low-rank structure, with the true rank $K = 2$ and $K = 10$, respectively. The small-world graph is not of an exact low-rank structure, and our method essentially provides a low-rank *approximation*. This scenario also allows us to investigate the performance of our method under potential model mis-specification. Besides, as none of the competitive solution enforces the symmetry property of the estimator $\widehat{\mathcal{B}}$, for comparison, we symmetrized their final estimators by $(\widehat{\mathcal{B}} + \widehat{\mathcal{B}}^\top)/2$.

Table 1: Third-order tensor response example with different sample size $n$, rank $K$, and sparsity $s$. Reported are the average estimation error, TPR, FPR, and F1 score, with the standard error shown in the parenthesis. The minimal error in each case is shown in boldface.

| n | K | s | Method | Error | TPR | FPR | F1 |
|---|---|---|---|---|---|---|---|
| 20 | 2 | 0.3 | OLS | 103.923 (2.330) | 1 (0) | 1 (0) | NA |
| | | | Sparse OLS | 57.208 (1.971) | 0.995 (0.001) | 0.085 (0.007) | 0.638 (0.019) |
| | | | ENV | 8.603 (0.142) | 1 (0) | 1 (0) | NA |
| | | | HOLRR | 7.311 (0.169) | 1 (0) | 1 (0) | NA |
| | | | STORE | **4.438** (0.098) | 1 (0) | 0 (0) | 1 (0) |
| | | 0.5 | OLS | 103.923 (2.330) | 1 (0) | 1 (0) | NA |
| | | | Sparse OLS | 76.709 (1.730) | 0.960 (0.004) | 0.372 (0.011) | 0.604 (0.005) |
| | | | ENV | 8.406 (0.109) | 1 (0) | 1 (0) | NA |
| | | | HOLRR | 7.278 (0.204) | 1 (0) | 1 (0) | NA |
| | | | STORE | **5.827** (0.101) | 1 (0) | 0 (0) | 1 (0) |
| | 5 | 0.3 | OLS | 103.923 (2.330) | 1 (0) | 1 (0) | NA |
| | | | Sparse OLS | 75.516 (1.319) | 0.996 (0) | 0.224 (0.041) | 0.657 (0.044) |
| | | | ENV | 13.593 (0.113) | 1 (0) | 1 (0) | NA |
| | | | HOLRR | 10.696 (0.307) | 1 (0) | 1 (0) | NA |
| | | | STORE | **7.086** (0.151) | 1 (0) | 0 (0) | 1 (0) |
| | | 0.5 | OLS | 103.923 (2.330) | 1 (0) | 1 (0) | NA |
| | | | Sparse OLS | 93.409 (2.038) | 0.988 (0.001) | 0.647 (0.008) | 0.738 (0.002) |
| | | | ENV | 13.625 (0.145) | 1 (0) | 1 (0) | NA |
| | | | HOLRR | 10.972 (0.294) | 1 (0) | 1 (0) | NA |
| | | | STORE | **9.140** (0.103) | 1 (0) | 0 (0) | 1 (0) |
| 100 | 2 | 0.3 | OLS | 44.250 (0.333) | 1 (0) | 1 (0) | NA |
| | | | Sparse OLS | 30.851 (1.626) | 1 (0) | 0.045 (0.020) | 0.831 (0.068) |
| | | | ENV | 3.721 (0.090) | 1 (0) | 1 (0) | NA |
| | | | HOLRR | 3.066 (0.068) | 1 (0) | 1 (0) | NA |
| | | | STORE | **1.979** (0.047) | 1 (0) | 0 (0) | 1 (0) |
| | | 0.5 | OLS | 44.250 (0.333) | 1 (0) | 1 (0) | NA |
| | | | Sparse OLS | 38.504 (0.443) | 0.993 (0) | 0.149 (0.003) | 0.801 (0.003) |
| | | | ENV | 3.803 (0.078) | 1 (0) | 1 (0) | NA |
| | | | HOLRR | 2.999 (0.063) | 1 (0) | 1 (0) | NA |
| | | | STORE | **2.609** (0.029) | 1 (0) | 0 (0) | 1 (0) |
| | 5 | 0.3 | OLS | 44.250 (0.333) | 1 (0) | 1 (0) | NA |
| | | | Sparse OLS | 32.584 (0.351) | 0.998 (0) | 0.149 (0.003) | 0.721 (0.004) |
| | | | ENV | 5.954 (0.049) | 1 (0) | 1 (0) | NA |
| | | | HOLRR | 4.335 (0.051) | 1 (0) | 1 (0) | NA |
| | | | STORE | **3.131** (0.033) | 1 (0) | 0 (0) | 1 (0) |
| | | 0.5 | OLS | 44.250 (0.333) | 1 (0) | 1 (0) | NA |
| | | | Sparse OLS | 41.487 (0.349) | 0.994 (0) | 0.459 (0.003) | 0.801 (0.001) |
| | | | ENV | 6.037 (0.027) | 1 (0) | 1 (0) | NA |
| | | | HOLRR | 4.328 (0.039) | 1 (0) | 1 (0) | NA |
| | | | STORE | **4.077** (0.037) | 1 (0) | 0 (0) | 1 (0) |

Table 2: Symmetric matrix response example with different graph structures. Reported are the average estimation error, TPR, FPR, and F1 score, with the standard error shown in the parenthesis. The minimal error in each example is shown in boldface.

| Graph Pattern | Method | Estimation Error | TPR | FPR | F1 |
|---|---|---|---|---|---|
| random | OLS | 22.168 (0.560) | 1 (0) | 1 (0) | NA |
| | Sparse OLS | 15.855 (0.543) | 1 (0) | 0.232 (0.012) | 0.646 (0.012) |
| | ENV | 10.751 (0.352) | 1 (0) | 1 (0) | NA |
| | HOLRR | 8.318 (0.199) | 1 (0) | 1 (0) | NA |
| | STORE | **2.497** (0.075) | 1 (0) | 0 (0) | 1 (0) |
| hub | OLS | 22.168 (0.560) | 1 (0) | 1 (0) | NA |
| | Sparse OLS | 12.924 (0.197) | 0.338 (0.060) | 0 (0) | 0.443 (0.073) |
| | ENV | 17.811 (0.222) | 1 (0) | 1 (0) | NA |
| | HOLRR | 11.685 (0.110) | 1 (0) | 1 (0) | NA |
| | STORE | **7.448** (0.453) | 0.824 (0.043) | 0.079 (0.004) | 0.272 (0.010) |
| small-world | OLS | 23.481 (0.635) | 1 (0) | 1 (0) | NA |
| | Sparse OLS | 22.024 (0.819) | 0.821 (0.063) | 0.010 (0.001) | 0.816 (0.062) |
| | ENV | 28.173 (0.427) | 1 (0) | 1 (0) | NA |
| | HOLRR | **18.794** (0.319) | 1 (0) | 1 (0) | NA |
| | STORE | 21.309 (0.154) | 0.901 (0.014) | 0.118 (0.004) | 0.610 (0.002) |

Table 2 reports the results over 20 data replications. Figure 2 further illustrates the estimated graph patterns of all methods in one replicate. The first column shows the true graphs. The second column shows the estimates from OLS, ENV, and HOLRR. Note that, although those three methods yield different numerical estimates, none of them impose any sparsity. As such, their estimated graph patterns look the same. The third column shows the estimate from Sparse OLS, and the last column from our method STORE. In the first scenario where the true graph is sparse and low-rank, our method substantially outperforms the competing solutions, with a much smaller estimation error and perfect selection accuracy. By contrast, OLS, ENV, and HOLRR yield much larger estimation error and could only generate dense graph estimates. Sparse OLS, on the other hand, produce many false positive links and a large error too. In the second scenario where the truth is a hub graph, our method again outperforms the alternative solutions. This time Sparse OLS misses many true positive links, which again results in a much larger estimation error. In the third scenario of the small-world graph, our model assumption is not satisfied. However our method still manages to achieve a reasonable estimation error and graph recovery. Although HOLRR obtains the smallest estimation error for this example, its graph estimate is completely dense and the corresponding FPR is one. Sparse OLS recoveres the graph pattern reasonably well, but still misses many true positive links, and yields a smaller TPR than our method.

Figure 2: Symmetric matrix response example. Shown are three graph patterns. Top row: the random graph, middle row: the hub graph, bottom row: the small-world graph. The first column: the true graph pattern; The second column: the estimated graph from OLS, ENV, and HOLRR; The third column: the estimated graph from Sparse OLS; The forth column: the estimated graph from our method STORE.

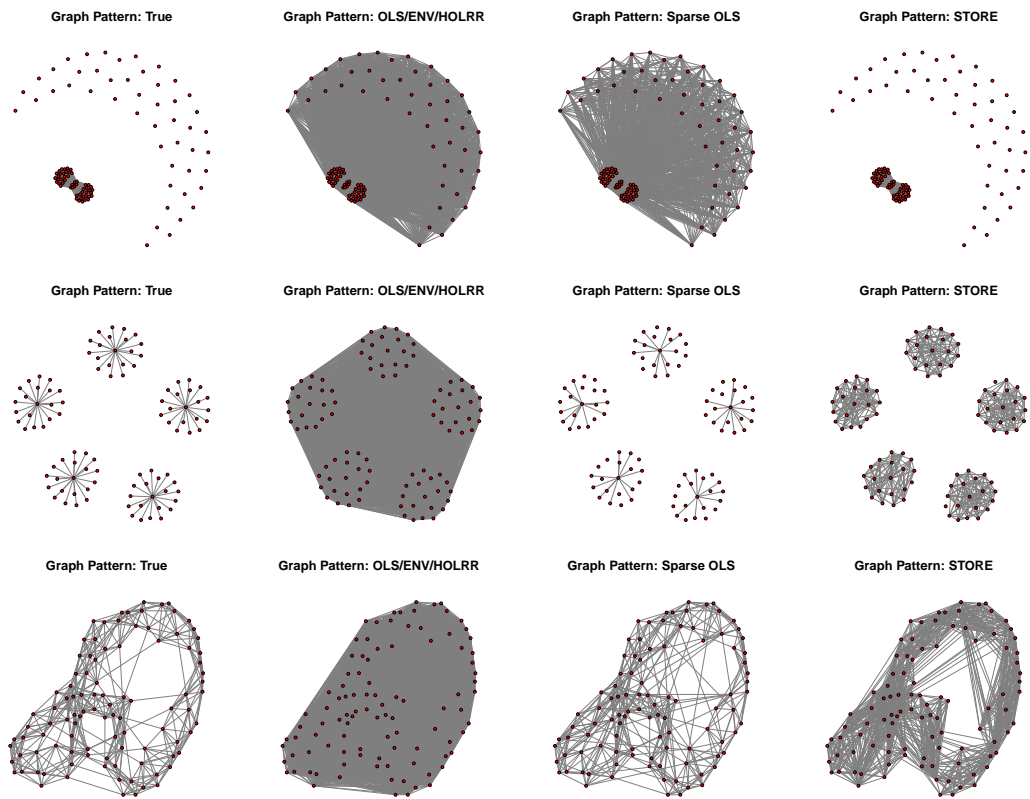

# 6 Real data analysis

In this section, we analyze a real neuroimaging data to illustrate our proposed method. The data is from the Autism Brain Imaging Data Exchange (ABIDE), a study for autism spectrum disorder (ASD) (Di Martino et al., 2014). ASD is an increasingly prevalent neurodevelopmental disorder. It is characterized by symptoms such as social difficulties, communication deficits, stereotyped behaviors and cognitive delays (Rudie et al., 2013). It is of scientific importance to understand how brain functional architecture and brain connectivity pattern differ between subjects with ASD and the normal controls. After removing those with missing values and poor image quality, the dataset consists of the resting-state functional magnetic resonance imaging (fMRI) of 795 subjects, of which 362 have ASD, and 433 are the normal controls. We took the fMRI image as the response, and the ASD status indicator (1 = ASD and 0 = normal control), age and sex as covariates. For each fMRI image, there are two forms of data summary: one is a 3D tensor and the other a 2D symmetric matrix. Correspondingly, we have fitted two separate tensor response regression models.

## 6.1 3D fALFF tensor response

The first form of the fMRI data is a third-order tensor of fractional amplitude of low-frequency fluctuations (fALFF). fALFF is a metric reflecting the percentage of power spectrum within low-frequency domain $(0.01 - 0.1$ Hz$)$. It characterizes the intensity of spontaneous brain activities, and provides a measure of functional architecture of the brain (Shi and Kang, 2015). It is calculated at each individual image voxel, and forms a 3D tensor with dimension $91 \times 109 \times 91$. We applied our STORE method as well as four competing methods (OLS, Sparse OLS, ENV, and HOLRR) to this general 3D tensor response.

Figure 3 shows the estimated coefficient tensors, overlaid on a brain image of a randomly selected subject, by all five methods. Among those, OLS, ENV, and HOLRR perform no region selection, whereas Sparse OLS yields a sparse estimate that is difficult to interpret. By contrast, the STORE estimate reveals a small number of brain regions that exhibit a clear difference between the ASD and normal control groups. Furthermore, we mapped those nonzero entries of our estimate to the AAL atlas. Table 3 reports the names of those identified regions and the corresponding coefficient entries in each region. Our results in general agree with the autism literature. For instance, we have found that multiple cerebellum (`Cerebellum`) regions show distinctive patterns between the ASD and control groups. The cerebellum has long been known for its importance in motor learning, coordination, and more recently, cognitive functions and affective regulation. It has emerged as one of the key brain regions affected in autism (Becker and Stoodley, 2013). Moreover, we identified the superior parietal lobule (`Parietal_Sup`) and precuneus (`Precuneus`), which agrees with Travers et al. (2015), in that they found individuals with ASD showed decreased activation in the superior parietal lobule and precuneus relative to individuals with typical development, suggesting that the superior parietal lobule may play an important role in motor learning and repetitive behavior in individuals with ASD.

Figure 3: Analysis of the ABIDE data. The response is a third-order brain image tensor. Shown are the estimated coefficient tensor overlaid on a randomly selected brain image. Top row: OLS/ENV/HOLRR, middle row: Sparse OLS, bottom row: STORE. Left column: front view, middle column: side view, right column: top view.

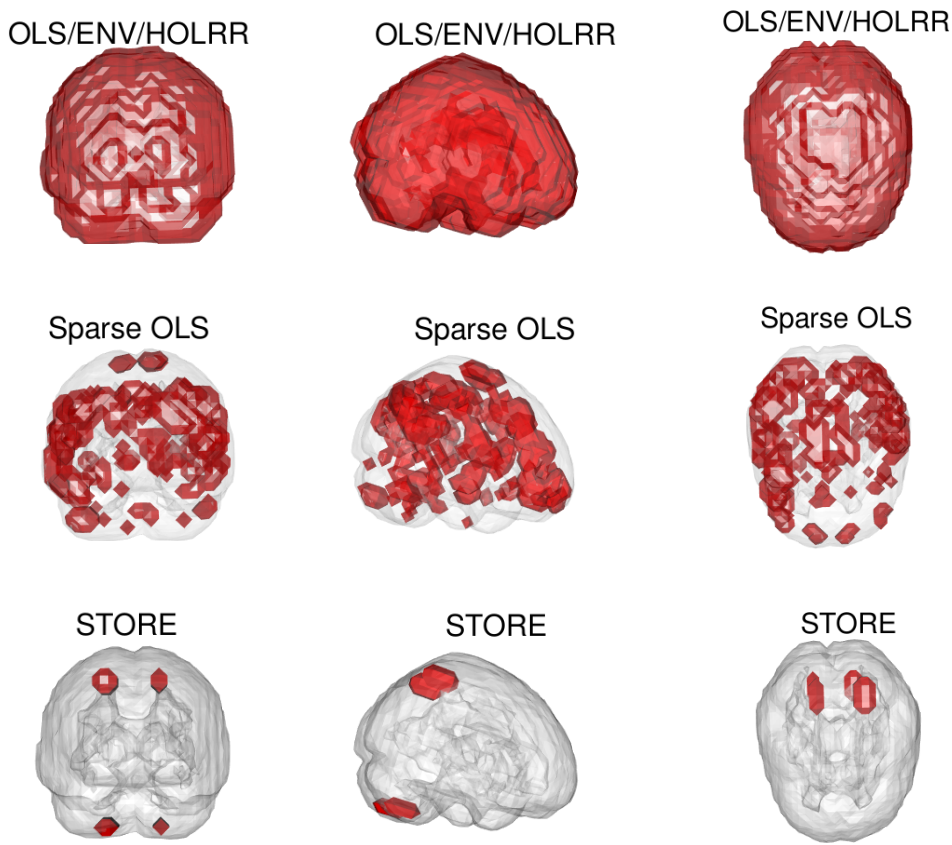

## 6.2 2D symmetric partial correlation matrix response

The second form of the fMRI data is a second-order symmetric partial correlation matrix. The partial correlation between brain regions reveals the synchronization of brain systems, and offers an alternative measure of the intrinsic brain functional architecture. It is evaluated, however, at a different scale from fALFF. Specifically, it is computed between pairs of pre-specified brain regions, each of which is a collection of brain voxels. It forms a 2D symmetric matrix with dimension $116 \times 116$, which corresponds to 116 regions-of-interest under the commonly used Anatomical Automatic Labeling (AAL) atlas (Tzourio-Mazoyer et al., 2002). We applied our STORE method to this 2D symmetric matrix response. As none of those competing methods incorporates the symmetry, we chose not to apply them in this analysis.

Table 3: Analysis of the ABIDE data. The response is the third-order fALFF tensor. Reported are the identified brain regions by our method that exhibit clear difference in fALFF between the ASD and normal control.

| Important Regions | $\widehat{\mathcal{B}}_{i,j,k}$ |
|---:|:---:|
| Postcentral_L | 0.0033 |
| Cerebelum_9_L | 0.0028 |
| Precuneus_L | 0.0025 |
| Occipital_Sup_R | 0.0025 |
| Parietal_Inf_L | 0.0025 |
| Cerebelum_8_L | 0.0020 |
| Cerebelum_8_R | 0.0020 |
| Precuneus_R | 0.0017 |
| Parietal_Sup_R | 0.0017 |
| Cerebelum_9_R | 0.0016 |
| Parietal_Sup_L | 0.0016 |

Figure 4 reports the top 20 identified links from our method. The red links correspond to those that are more likely to be absent in the ASD group, whereas the black ones are those more likely to be present in ASD. Table 4 reports the names of those links and their relative connection strengthens in the network, as reflected by the corresponding entries in the matrix coefficient $\widehat{\mathcal{B}}$. Many different connectivity patterns concentrate on the left middle frontal gyrus (Frontal_Mid_L) and the temporal lobe (Temporal), and such findings again agrees with the autism literature (Kana et al., 2006; Di Martino et al., 2014; Ha et al., 2015).

Figure 4: Analysis of the ABIDE data. The response is a symmetric partial correlation matrix. Shown are the top 20 links found by our method. Red links are more likely to be absent in the ASD group and black link are more likely to be present in the ASD group.

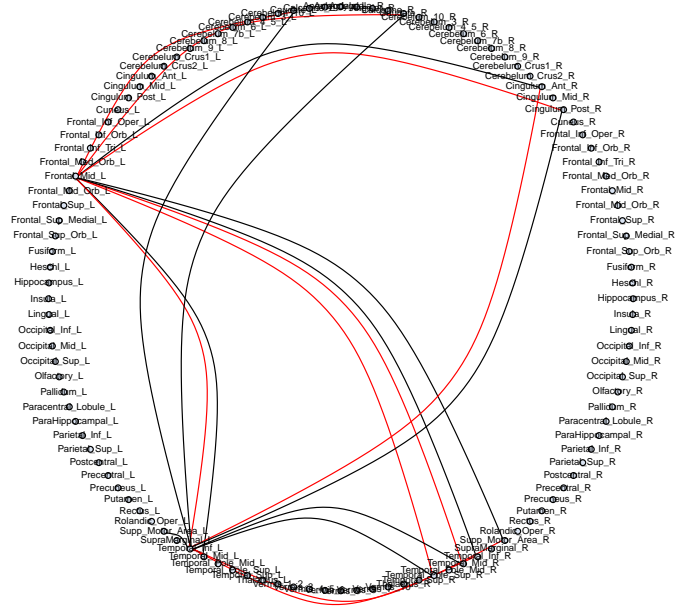

Table 4: Analysis of the ABIDE data. The response is a symmetric partial correlation matrix. Reported are the identified top 20 links by our method.

| Connected Regions | $\widehat{\mathcal{B}}_{i,j}$ |
|---|---|
| Frontal_Mid_L ⟷ Temporal_Inf_L | -0.0034 |
| Cingulum_Ant_R ⟷ Temporal_Inf_L | -0.0027 |
| Temporal_Inf_L ⟷ Temporal_Inf_R | -0.0024 |
| Supp_Motor_Area_R ⟷ Temporal_Inf_L | -0.0023 |
| Temporal_Mid_L ⟷ Temporal_Inf_L | -0.0023 |
| Frontal_Mid_L ⟷ Temporal_Mid_R | -0.0020 |
| Frontal_Mid_L ⟷ Caudate_R | -0.0019 |
| Frontal_Mid_L ⟷ Temporal_Pole_Sup_R | -0.0018 |
| Frontal_Mid_L ⟷ Cingulum_Post_R | -0.0017 |
| Frontal_Mid_L ⟷ Caudate_L | -0.0016 |
| Frontal_Mid_L ⟷ Supp_Motor_Area_R | 0.0017 |
| Frontal_Mid_L ⟷ Temporal_Mid_L | 0.0017 |
| Frontal_Mid_L ⟷ Temporal_Inf_R | 0.0019 |
| Frontal_Mid_L ⟷ Cingulum_Ant_R | 0.0021 |
| Supp_Motor_Area_L ⟷ Temporal_Inf_L | 0.0021 |
| Caudate_L ⟷ Temporal_Inf_L | 0.0021 |
| Cingulum_Post_R ⟷ Temporal_Inf_L | 0.0023 |
| Temporal_Pole_Sup_R ⟷ Temporal_Inf_L | 0.0024 |
| Caudate_R ⟷ Temporal_Inf_L | 0.0026 |
| Temporal_Mid_R ⟷ Temporal_Inf_L | 0.0026 |

# Appendix

We divide the appendix into three parts. First we begin with a number of auxiliary lemmas. Next we provide detailed proofs of the main theorem. Then we present the technical proofs of all lemmas and corollaries.

## A1 Auxiliary lemmas

**Lemma 3.** *For any matrix* $\mathbf{Y}_i \in \mathbb{R}^{d_1 \times d_2}$, *with* $i = 1, \ldots, n$, *and vectors* $\gamma = (\gamma_1, \ldots, \gamma_n)^\top \in \mathbb{R}^n$, $\alpha \in \mathbb{R}^{d_1}$ *and* $\beta \in \mathbb{R}^{d_2}$, *we have*

$$\arg\min_{\beta, \|\beta\|_2=1} \frac{1}{n} \sum_{i=1}^n \gamma_i^2 \left\| \mathbf{Y}_i - \alpha\beta^\top \right\|_F^2 = \arg\min_{\beta, \|\beta\|_2=1} \left\| \frac{1}{n} \sum_{i=1}^n \gamma_i^2 \mathbf{Y}_i - \alpha\beta^\top \right\|_F^2.$$

The proof of Lemma 3 is provided in Section A3.3 in the online supplement.

We next introduce the Slepian's lemma (Slepian, 1962), which provides a comparison between the supremums of two Gaussian processes.

**Lemma 4.** *(Slepian, 1962, Slepian's Lemma) Denote two centered Gaussian processes $\{G_s, s \in \mathcal{S}\}$ and $\{H_s, s \in \mathcal{S}\}$. Assume that both processes are almost surely bounded and for each $s, t \in \mathcal{S}$, $\mathbb{E}(G_s - G_t)^2 \leq E(H_s - H_t)^2$, then we have*

$$\mathbb{E}\left[\sup_{s \in \mathcal{S}} G_s\right] \leq \mathbb{E}\left[\sup_{s \in \mathcal{S}} H_s\right].$$

*Moreover, if $\mathbb{E}(G_s^2) = E(H_s^2)$ for all $s \in \mathcal{S}$, then we have, for each $x > 0$,*

$$\mathbb{P}\left[\sup_{s \in \mathcal{S}} G_s > x\right] \leq \mathbb{P}\left[\sup_{s \in \mathcal{S}} H_s > x\right].$$

The next result provides a concentration of Lipschitz functions of Gaussian random variables (Massart, 2003).

**Lemma 5.** *(Massart, 2003, Theorem 3.4) Let $\mathbf{v} \in \mathbb{R}^d$ be a Gaussian random variable such that $\mathbf{v} \sim N(0, \mathbf{I}_d)$. Assuming $g(\mathbf{v}) \in \mathbb{R}$ to be a Lipschitz function such that $|g(\mathbf{v}_1) - g(\mathbf{v}_2)| \leq L\|\mathbf{v}_1 - \mathbf{v}_2\|_2$ for any $\mathbf{v}_1, \mathbf{v}_2 \in \mathbb{R}^d$, then we have, for each $t > 0$,*

$$\mathbb{P}\left[|g(\mathbf{v}) - \mathbb{E}[g(\mathbf{v})]| \geq t\right] \leq 2 \exp\left(-\frac{t^2}{2L^2}\right).$$

The next lemma provides an upper bound of the Gaussian width of the unit ball for the sparsity regularizer (Raskutti and Yuan, 2016).

**Lemma 6.** *(Raskutti and Yuan, 2016, Lemma 2) For a tensor $\mathcal{T} \in \mathbb{R}^{d_1 \times d_2 \times d_3}$, denote its regularizer $R(\mathcal{T}) = \sum_{j_1} \sum_{j_2} \sum_{j_3} |\mathcal{T}_{j_1, j_2, j_3}|$. Define the unit ball of this regularizer as $B_R(1) := \{\mathcal{T} \in \mathbb{R}^{d_1 \times d_2 \times d_3} | R(\mathcal{T}) \leq 1\}$. For a Gaussian tensor $\mathcal{G} \in \mathbb{R}^{d_1 \times d_2 \times d_3}$ whose entries are independent standard normal random variables, we have*

$$\mathbb{E}\left[\sup_{\mathcal{T} \in B_R(1)} \langle \mathcal{T}, \mathcal{G}\rangle\right] \leq c\sqrt{\log(d_1 d_2 d_3)},$$

*for some bounded constant $c > 0$.*

The next lemma links the hard thresholding sparsity and the $L_1$-penalized sparsity.

**Lemma 7.** *For any vectors $\mathbf{u} \in \mathbb{R}^{d_1}, \mathbf{v} \in \mathbb{R}^{d_2}, \mathbf{w} \in \mathbb{R}^{d_3}$ satisfying $\|\mathbf{u}\|_2 = \|\mathbf{v}\|_2 = \|\mathbf{w}\|_2 = 1, \|\mathbf{u}\|_0 \leq s, \|\mathbf{v}\|_0 \leq s$, and $\|\mathbf{w}\|_0 \leq s$, denoting $\mathcal{A} := \mathbf{u} \circ \mathbf{v} \circ \mathbf{w}$, we have the bound of the $L_1$-norm regularizer*

$$\|\mathcal{A}\|_1 := \sum_{j_1} \sum_{j_2} \sum_{j_3} |\mathcal{A}_{j_1 j_2 j_3}| \leq s^{3/2}.$$

## A2 Proof of Theorem 1

We divide the proof of Theorem 1 into two major steps: characterization of the estimation error in Step 1 in Algorithm 1, then the estimation error in Step 2. Each leads to a new theorem. Then we complete the proof of Theorem 1 by iteratively applying those results.

### A2.1 Estimation error in Step 1 of Algorithm 1

We first derive the estimation error in Step 1 of our Algorithm 1. The key idea is to transform the problem into a standard sparse tensor decomposition problem, then incorporate the existing contracting results obtained in Sun et al. (2017) to derive the final error bound of the estimator in Step 1. In the following derivation, for simplicity, we assume $K = 1$. This does not lose generality, since $K$ is assumed to be a constant, and it does not affect the final error rate. In this case, the true model reduces to $\mathcal{Y}_i = w^*(\boldsymbol{\beta}_{k,m+1}^{*\top}\mathbf{x}_i)\boldsymbol{\beta}_{k,1}^* \circ \cdots \circ \boldsymbol{\beta}_{k,m}^* + \mathcal{E}_i$, for each $i = 1, \ldots, n$. Based on the Step 1 of our algorithm, if the true parameter $\boldsymbol{\beta}_{k,m+1}^*$ is available, we are solving the following sparse tensor decomposition problem,

$$\bar{\mathcal{R}}_k = \mathcal{T} + \bar{\mathcal{E}}, \tag{A1}$$

where the true tensor $\mathcal{T} = w_k^*\boldsymbol{\beta}_{k,1}^* \circ \cdots \circ \boldsymbol{\beta}_{k,m}^*$, the oracle response and the oracle error are, respectively,

$$\bar{\mathcal{R}}_k := \frac{1}{n}\sum_{i=1}^n \mathcal{R}_{ik} = \frac{1}{n}\sum_{i=1}^n \frac{\mathcal{Y}_i}{\boldsymbol{\beta}_{k,m+1}^{*\top}\mathbf{x}_i} \text{ and } \bar{\mathcal{E}} = \frac{1}{n}\sum_{i=1}^n \frac{\mathcal{E}_i}{\boldsymbol{\beta}_{k,m+1}^{*\top}\mathbf{x}_i}.$$

In practice, however, we only have an estimator $\widehat{\boldsymbol{\beta}}_{k,m+1}$. Hence the sparse tensor decomposition method in Step 1 of our algorithm is actually applied to

$$\widehat{\mathcal{R}}_k = \mathcal{T} + \widehat{\mathcal{E}} \tag{A2}$$

with the response tensor

$$\widehat{\mathcal{R}}_k := \frac{1}{n}\sum_{i=1}^n \frac{\mathcal{Y}_i}{\widehat{\boldsymbol{\beta}}_{k,m+1}^{\top}\mathbf{x}_i}$$

Therefore, according to ($A1$) and ($A2$), we have the explicit form of $\widehat{\mathcal{E}}$,

$$\widehat{\mathcal{E}} = \widehat{\mathcal{R}}_k - \bar{\mathcal{R}}_k + \bar{\mathcal{E}} \\ = \underbrace{\frac{1}{n}\sum_{i=1}^{n}\frac{(\boldsymbol{\beta}^*_{k,m+1} - \widehat{\boldsymbol{\beta}}_{k,m+1})^\top \mathbf{x}_i}{\widehat{\boldsymbol{\beta}}^\top_{k,m+1}\mathbf{x}_i \boldsymbol{\beta}^{*\top}_{k,m+1}\mathbf{x}_i}\mathcal{Y}_i}_{I_1} + \underbrace{\frac{1}{n}\sum_{i=1}^{n}\frac{\mathcal{E}_i}{\boldsymbol{\beta}^{*\top}_{k,m+1}\mathbf{x}_i}}_{I_2} \quad (A3)$$

Before we derive the estimation error of the estimator based on ($A2$), we introduce a lemma for deriving the error bound of a general sparse tensor decomposition.

**Assumption 4.** *The decomposition components are incoherent such that*

$$\zeta := \max_{i \neq j}\{|\langle\boldsymbol{\beta}^*_{i,1},\boldsymbol{\beta}^*_{j,1}\rangle|, |\langle\boldsymbol{\beta}^*_{i,2},\boldsymbol{\beta}^*_{j,2}\rangle|, |\langle\boldsymbol{\beta}^*_{i,3},\boldsymbol{\beta}^*_{j,3}\rangle|\} \leq \frac{C_0}{\sqrt{d_0}},$$

*with $d_0 = \max\{d_{01}, d_{02}, d_{03}\}$, and for any $j$, $\|\sum_{i \neq j} w_i \langle\boldsymbol{\beta}^*_{i,1},\boldsymbol{\beta}^*_{j,1}\rangle\langle\boldsymbol{\beta}^*_{i,2},\boldsymbol{\beta}^*_{j,2}\rangle\boldsymbol{\beta}^*_{i,3}\| \leq C_1 w^*_{\max}\sqrt{K}\zeta$. Moreover, the matrices $\mathbf{A} := [\boldsymbol{\beta}^*_{i,1},\cdots,\boldsymbol{\beta}^*_{K,1}]$, $\mathbf{B} := [\boldsymbol{\beta}^*_{i,2},\cdots,\boldsymbol{\beta}^*_{K,2}]$, and $\mathbf{C} := [\boldsymbol{\beta}^*_{i,3},\cdots,\boldsymbol{\beta}^*_{K,3}]$ satisfy that $\max\{\|\mathbf{A}\|,\|\mathbf{B}\|,\|\mathbf{C}\|\} \leq 1 + C_2\sqrt{K/d_0}$ for some positive constants $C_0, C_1, C_2$.*

Define a function $f(\epsilon; K, d_0)$ as

$$f(\epsilon; K, d_0) := \frac{2C_0}{\sqrt{d_0}}\left(1 + C_2\sqrt{\frac{K}{d_0}}\right)^2 \epsilon + C_1\frac{\sqrt{K}}{d_0} + C_3\epsilon^2,$$

for some constants $C_0, C_1, C_2, C_3 > 0$. When $K = o(d_0^{3/2})$, the first two terms of $f(\epsilon; K, d_0)$ converge to 0 and the last term is the contracting term.

**Lemma 8.** *(Sun et al., 2017, Lemma S.4.1) Consider the model $\widehat{\mathcal{T}} = \mathcal{T} + \mathcal{E}$ where the low-rank and sparse components of $\mathcal{T}$ satisfy Assumption 4, and assume $\|\mathcal{T}\| \leq C_3 w^*_{\max}$ and $K = o(d_0^{3/2})$. In addition, assume the estimators $\widehat{\boldsymbol{\beta}}_{j,1}$ and $\widehat{\boldsymbol{\beta}}_{j,2}$ satisfy $D(\widehat{\boldsymbol{\beta}}_{j,1},\boldsymbol{\beta}^*_{j,1}) \leq \epsilon$ and $D(\widehat{\boldsymbol{\beta}}_{j,2},\boldsymbol{\beta}^*_{j,2}) \leq \epsilon$ for some $j \in [K]$. If the perturbation error $\eta(\mathcal{E}, d_0 + s)$, with $s \geq d_0$, is small enough such that $\eta(\mathcal{E}, d_0 + s) < w_j(1 - \epsilon^2) - w^*_{\max}f(\epsilon; K, d_0)$, then the update $\widehat{\boldsymbol{\beta}}_{j,3}$ satisfies, with high probability,*

$$D(\widehat{\boldsymbol{\beta}}_{j,3},\boldsymbol{\beta}^*_{j,3}) \leq \frac{\sqrt{5}w^*_{\max}f(\epsilon; K, d_0) + \sqrt{5}\eta(\mathcal{E}, d_0 + s)}{w_j(1 - \epsilon^2) - w^*_{\max}f(\epsilon; K, d_0) - \eta(\mathcal{E}, d_0 + s)}.$$

*If we further assume $D(\widehat{\boldsymbol{\beta}}_{j,3},\boldsymbol{\beta}^*_{j,3}) \leq \epsilon$, then the update $\widehat{w} = \widehat{\mathcal{T}} \times_1 \widehat{\boldsymbol{\beta}}_{j,1} \times_2 \widehat{\boldsymbol{\beta}}_{j,2} \times_3 \widehat{\boldsymbol{\beta}}_{j,3}$ satisfies, with high probability, $|\widehat{w} - w_j| \leq 2w_j\epsilon^2 + w^*_{\max}f(\epsilon; K, d_0) + \eta(\mathcal{E}, d_0 + s)$.*

By verifying the conditions in Lemma 8, we are able to compute the estimation error in Step 1 of our algorithm. For simplicity, we consider the case when $m = 3$, while the extension to a more general $m$ is straightforward.

**Lemma 9.** *Assume $\|\mathcal{T}\| \leq C_1 w^*_{\max}$, $D(\widehat{\boldsymbol{\beta}}_{k,1}, \boldsymbol{\beta}^*_{k,1}) \leq \epsilon$, $D(\widehat{\boldsymbol{\beta}}_{k,2}, \boldsymbol{\beta}^*_{k,2}) \leq \epsilon$, and*

$$\epsilon < \min\left\{\sqrt{\frac{w^*_{\min}}{2(w^*_{\min} + w^*_{\max}C_1)}}, \frac{w^*_{\min}}{4\sqrt{5}w^*_{\max}C_1}\right\}.$$

*Assume the error tensor satisfies $\eta(\widehat{\mathcal{E}}, d_0 + s) \leq w^*_{\min}/4$ with $s \geq d_0$, where $\widehat{\mathcal{E}}$ is as defined in (A3). Then we have*

$$D(\widehat{\boldsymbol{\beta}}_{k,3}, \boldsymbol{\beta}^*_{k,3}) \leq \kappa_1 \epsilon + \frac{4\sqrt{5}}{w^*_{\min}}\eta(\widehat{\mathcal{E}}, d_0 + s),$$

*where the contraction coefficient*

$$\kappa_1 := \frac{4\sqrt{5}w^*_{\max}C_1}{w^*_{\min}}\epsilon < \frac{4\sqrt{5}w^*_{\max}C_1}{w^*_{\min}}\min\left\{\sqrt{\frac{w^*_{\min}}{2(w^*_{\min} + w^*_{\max}C_1)}}, \frac{w^*_{\min}}{4\sqrt{5}w^*_{\max}C_1}\right\} \in (0, 1).$$

The proof of Lemma 9 is provided in Section A3.5. Based on Lemma 9, to compute the closed-form error rate in Step 1 of our Algorithm, the remaining step is to compute $\eta(\widehat{\mathcal{E}}, s)$, since $\eta(\widehat{\mathcal{E}}, d_0 + s) \leq 2\eta(\widehat{\mathcal{E}}, s)$ by noting that $s \geq d_0$. Here the explicit form of $\widehat{\mathcal{E}}$ is defined in (A3). Again we only consider $m = 3$ for simplicity, and the proof for a general $m$ follows straightforwardly.

**Lemma 10.** *Assume the conditions in Lemma 9 hold. Assume that $\|w^*_k \boldsymbol{\beta}^*_{k,1} \circ \cdots \circ \boldsymbol{\beta}^*_{k,m}\| \leq C_1$, $\|\mathbf{x}_i\| \leq C_2$, and $|\boldsymbol{\beta}^{*\top}_{k,m+1}\mathbf{x}_i| \geq C_3$ for each $i = 1, \ldots, n$, for some positive constants $C_1, C_2, C_3$. If $\epsilon \leq C_3/(2C_2)$, then we have*

$$\eta(\widehat{\mathcal{E}}, s) \leq \underbrace{\frac{2C_2\eta(\bar{\mathcal{E}}, s)}{C_3^2}}_{\kappa_2}\epsilon + \frac{1}{C_3}\eta(\bar{\mathcal{E}}, s).$$

*where $\bar{\mathcal{E}} := \frac{1}{n}\sum_{i=1}^n \mathcal{E}_i$.*

The proof of Lemma 10 is provided in Section A3.6. Combining Lemma 9 and Lemma 10, we obtain the final contraction result of Step 1 in Algorithm 1.

**Theorem 2.** *(Contraction result in Step 1 in Algorithm 1)* Assume $D(\widehat{\boldsymbol{\beta}}_{k,1}, \boldsymbol{\beta}^*_{k,1}) \leq \epsilon$, $D(\widehat{\boldsymbol{\beta}}_{k,2}, \boldsymbol{\beta}^*_{k,2}) \leq \epsilon$, and $D(\widehat{\boldsymbol{\beta}}_{k,4}, \boldsymbol{\beta}^*_{k,4}) \leq \epsilon$, with

$$\epsilon < \min\left\{\sqrt{\frac{w^*_{\min}}{2(w^*_{\min} + w^*_{\max}C_1)}}, \frac{w^*_{\min}}{4\sqrt{5}w^*_{\max}C_1}, \frac{C_3}{2C_2}\right\}.$$

*Assume the error tensor satisfies $\eta(n^{-1}\sum_{i=1}^n \mathcal{E}_i, d_0 + s) \leq w^*_{\min}/4$ with $s \geq d_0$. Then we have*

$$D(\widehat{\boldsymbol{\beta}}_{k,3}, \boldsymbol{\beta}^*_{k,3}) \leq (\kappa_1 + \kappa_2)\epsilon + \frac{4\sqrt{5}}{C_3 w^*_{\min}}\eta\left(\frac{1}{n}\sum_{i=1}^n \mathcal{E}_i, s\right).$$

### A2.2 Estimation error in Step 2 of Algorithm 1

Next we derive the estimation error in Step 2 of our algorithm. That is, we aim to bound $D(\widehat{\boldsymbol{\beta}}_{k,m+1}, \boldsymbol{\beta}^*_{k,m+1})$ given the estimators $\widehat{w}_k, \widehat{\boldsymbol{\beta}}_{k,1}, \ldots, \widehat{\boldsymbol{\beta}}_{k,m}$.

Denote $\widehat{\mathcal{A}}_k = \widehat{w}_k \widehat{\boldsymbol{\beta}}_{k,1} \circ \cdots \circ \widehat{\boldsymbol{\beta}}_{k,m}$, and $\widehat{\mathcal{T}}_i = \mathcal{Y}_i - \sum_{k' \neq k, k' \in [K]} \widehat{w}_{k'} (\widehat{\boldsymbol{\beta}}^\top_{k',m+1}\mathbf{x}_i)\widehat{\boldsymbol{\beta}}_{k',1} \circ \cdots \circ \widehat{\boldsymbol{\beta}}_{k',m}$, and the closed-form estimator in Step 2 of our algorithm is

$$\widehat{\boldsymbol{\beta}}_{k,m+1} = \left(\frac{1}{n}\sum_{i=1}^n \mathbf{x}_i\mathbf{x}_i^\top\right)^{-1} \frac{n^{-1}\sum_{i=1}^n \langle \widehat{\mathcal{T}}_i, \widehat{\mathcal{A}}_k\rangle \mathbf{x}_i}{\|\widehat{\mathcal{A}}_k\|_F^2}.$$

**Theorem 3.** *(Contraction result in Step 2 in Algorithm 1)* Under Assumption 1, and the assumption that the initialization error satisfies $\epsilon \leq w^*_{\min}/2$, if $|\widehat{w}_k - w^*_k| \leq \epsilon$, $D(\widehat{\boldsymbol{\beta}}_{k,1}, \boldsymbol{\beta}^*_{k,1}) \leq \epsilon, \ldots, D(\widehat{\boldsymbol{\beta}}_{k,m}, \boldsymbol{\beta}^*_{k,m}) \leq \epsilon$, then we have

$$D(\widehat{\boldsymbol{\beta}}_{k,m+1}, \boldsymbol{\beta}^*_{k,m+1}) \leq \kappa_3 \epsilon + \frac{\widetilde{C}}{\sqrt{n}},$$

*where $\kappa_3 := 2/w^*_{\min} + 6\sqrt{2}$ and $\widetilde{C}$ is a positive constant as defined in (A7).*

*Proof*: For simplicity, we only prove for $K = 1$ and $m = 3$. The derivation for a general $K$ and $m$ follows similarly.

Denote $\mathcal{A}^*_k := w^*_k \boldsymbol{\beta}^*_{k,1} \circ \boldsymbol{\beta}^*_{k,2} \circ \boldsymbol{\beta}^*_{k,3}$. The true model reduces to $\mathcal{Y}_i = (\boldsymbol{\beta}^{*\top}_{k,4}\mathbf{x}_i)\mathcal{A}^*_k + \mathcal{E}_i$, for each $i = 1, \ldots, n$. Denote $\Omega := (n^{-1}\sum_{i=1}^n \mathbf{x}_i\mathbf{x}_i^\top)^{-1}$. Given $\widehat{\mathcal{A}}_k = \widehat{w}_k \widehat{\boldsymbol{\beta}}_{k,1} \circ \cdots \circ \widehat{\boldsymbol{\beta}}_{k,3}$, and $\widehat{\mathcal{T}}_i = \mathcal{Y}_i$ when $K = 1$, we have the following simplification of $\widehat{\boldsymbol{\beta}}_{k,4}$,

$$\begin{aligned}\widehat{\boldsymbol{\beta}}_{k,4} &= \Omega \frac{\sum_{i=1}^n \langle \widehat{\mathcal{T}}_i, \widehat{\mathcal{A}}_k\rangle \mathbf{x}_i}{n\|\widehat{\mathcal{A}}_k\|_F^2} = \Omega \frac{\sum_{i=1}^n \langle (\boldsymbol{\beta}^{*\top}_{k,4}\mathbf{x}_i)\mathcal{A}^*_k + \mathcal{E}_i, \widehat{\mathcal{A}}_k\rangle \mathbf{x}_i}{n\|\widehat{\mathcal{A}}_k\|_F^2} \\ &= \frac{\langle \mathcal{A}^*_k, \widehat{\mathcal{A}}_k\rangle}{\|\widehat{\mathcal{A}}_k\|_F^2}\boldsymbol{\beta}^*_{k,4} + \frac{\Omega \sum_{i=1}^n \langle \mathcal{E}_i, \widehat{\mathcal{A}}_k\rangle \mathbf{x}_i}{n\|\widehat{\mathcal{A}}_k\|_F^2},\end{aligned} \quad (A4)$$

where the first part in $(A4)$ is due to the fact that $n^{-1}\sum_{i=1}^n (\boldsymbol{\beta}_{k,4}^{*\top}\mathbf{x}_i)\Omega\mathbf{x}_i = n^{-1}\sum_{i=1}^n \Omega\mathbf{x}_i\mathbf{x}_i^\top \boldsymbol{\beta}_{k,4}^* = \boldsymbol{\beta}_{k,4}^*$. Therefore, the error bound of $\widehat{\boldsymbol{\beta}}_{k,4}$ can be simplified as

$$\left\|\widehat{\boldsymbol{\beta}}_{k,4} - \boldsymbol{\beta}_{k,4}^*\right\|_2 = \left\|\frac{\langle \mathcal{A}_k^*, \widehat{\mathcal{A}}_k\rangle}{\|\widehat{\mathcal{A}}_k\|_F^2}\boldsymbol{\beta}_{k,4}^* - \boldsymbol{\beta}_{k,4}^* + \frac{\Omega\sum_{i=1}^n \langle \mathcal{E}_i, \widehat{\mathcal{A}}_k\rangle \mathbf{x}_i}{n\|\widehat{\mathcal{A}}_k\|_F^2}\right\|_2,$$

$$\leq \underbrace{\left|\frac{\langle \mathcal{A}_k^*, \widehat{\mathcal{A}}_k\rangle - \|\widehat{\mathcal{A}}_k\|_F^2}{\|\widehat{\mathcal{A}}_k\|_F^2}\right|}_{(I)} + \underbrace{\frac{\|\Omega\sum_{i=1}^n\langle\mathcal{E}_i,\widehat{\mathcal{A}}_k\rangle\mathbf{x}_i\|_2}{n\|\widehat{\mathcal{A}}_k\|_F^2}}_{(II)}. \quad (A5)$$

In the following lemma, we bound the two terms in $(A5)$ to obtain the final error bound.

**Lemma 11.** *Under the Conditions in Theorem 3, we have*

$$(I) \leq \left[\frac{2}{w_{\min}^*} + 6\sqrt{2}\right]\epsilon, \quad (A6)$$

$$(II) \leq \underbrace{\frac{\max_i\|\Omega\mathbf{x}_i\|_2 \cdot \mathbb{E}\left[\|\mathcal{G}\|_F\right]}{\|\widehat{\mathcal{A}}_k\|_F}}_{\widetilde{C}} \cdot \frac{1}{\sqrt{n}}. \quad (A7)$$

Finally, combining the results in $(A6)$ and $(A7)$ leads to the final bound of $\|\widehat{\boldsymbol{\beta}}_{k,4} - \boldsymbol{\beta}_{k,4}^*\|_2$, and hence the bound of $D(\widehat{\boldsymbol{\beta}}_{k,4}, \boldsymbol{\beta}_{k,4}^*)$. □

### A2.3 Proof of Theorem 1

Now we complete the proof of Theorem 1, by iteratively applying the contraction results in Theorems 2 and 3 for the two steps of Algorithm 1. In iteration $t = 1$, given the initializations $\widehat{\boldsymbol{\beta}}_{k,j}^{(0)}$ and $\widehat{w}^{(0)}$ with initialization error $\epsilon$, Theorem 2 implies that

$$D(\widehat{\boldsymbol{\beta}}_{k,3}^{(1)}, \boldsymbol{\beta}_{k,3}^*) \leq (\kappa_1 + \kappa_2)\epsilon + \frac{4\sqrt{5}}{C_3 w_{\min}^*}\eta\left(\frac{1}{n}\sum_{i=1}^n \mathcal{E}_i, s\right),$$

where $\kappa_1 + \kappa_2 < 1$ according to Assumptions 2 and 3. The second term converges to zero as sample size increases. Therefore, for a sufficiently large sample size, the above error bound is smaller than $\epsilon$. By a similar derivation, the same error bound holds for $D(\widehat{\boldsymbol{\beta}}_{k,1}^{(1)}, \boldsymbol{\beta}_{k,1}^*)$, $D(\widehat{\boldsymbol{\beta}}_{k,2}^{(1)}, \boldsymbol{\beta}_{k,2}^*)$, and $|\widehat{w}_k^{(1)} - w_k^*|$. In Step 2 of the algorithm, applying Theorem 3 based on the above estimators in iteration $t = 1$, we obtain that

$$D(\widehat{\boldsymbol{\beta}}_{k,4}^{(1)}, \boldsymbol{\beta}_{k,4}^*) \leq \kappa_3(\kappa_1 + \kappa_2)\epsilon + \kappa_3\frac{4\sqrt{5}}{C_3 w_{\min}^*}\eta\left(\frac{1}{n}\sum_{i=1}^n \mathcal{E}_i, s\right) + \frac{\widetilde{C}}{\sqrt{n}}.$$

Again, the contraction coefficient $\kappa_3(\kappa_1 + \kappa_2) < 1$ according to Assumptions 2 and 3, and the remaining term converges to zero as sample size increases. For $\kappa = (\kappa_1 + \kappa_2)\kappa_3 \in (0, 1)$, by repeatedly applying these derivations, in the $t$ iteration, we obtain that

$$\max \left\{ \max_k \|\widehat{w}_k^{(t)} - w_k^*\|_2, \max_{k,j} \{\|\widehat{\boldsymbol{\beta}}_{k,j}^{(t)} - \boldsymbol{\beta}_{k,j}^*\|_2\} \right\}$$
$$\leq \kappa^t \epsilon + \frac{1-\kappa^t}{1-\kappa} \frac{4\sqrt{5}}{C_3 w_{\min}^*} \eta \left( \frac{1}{n} \sum_{i=1}^n \mathcal{E}_i, s \right) + \frac{1-\kappa^{t-1}}{1-\kappa} \frac{\widetilde{C}}{\sqrt{n}}$$
$$\leq \kappa^t \epsilon + \frac{1}{1-\kappa} \max \left\{ \frac{4\sqrt{5}}{C_3 w_{\min}^*} \eta \left( \frac{1}{n} \sum_{i=1}^n \mathcal{E}_i, s \right), \frac{\widetilde{C}}{\sqrt{n}} \right\}.$$

This completes the proof of Theorem 1. $\square$

## A3 Proofs of lemmas and corollaries

### A3.1 Proof of Lemma 1

To solve (4), we can use the alternating updating method to update one parameter at a time. In particular, for each $j = 1, \ldots, m$, given $\beta_k^{(j')}$ with $j' \neq j$, we solve

$$\widehat{\beta}_{k,j} := \arg \min_{\substack{\beta_{k,j} \\ \|\beta_{k,j}\|_2=1, \|\beta_{k,j}\|_0 \leq s_j}} \frac{1}{n} \sum_{i=1}^n \alpha_{ik}^2 \left\| \mathcal{R}_i - w_k \beta_{k,1} \circ \cdots \circ \beta_{k,m} \right\|_F^2.$$

According to the matrix representation of tensor operations (Kolda and Bader, 2009; Kim et al., 2014), this optimization problem is equivalent to solve

$$\min_{\substack{\beta_{k,j} \\ \|\beta_{k,j}\|_2=1, \|\beta_{k,j}\|_0 \leq s_j}} \frac{1}{n} \sum_{i=1}^n \alpha_{ik}^2 \left\| [\mathcal{R}_i]_{(j)}^\top - \mathbf{h}_k^{(j)} \beta_k^{(j)\top} \right\|_F^2, \tag{A8}$$

where $[\mathcal{R}_i]_{(j)}$ is the mode-$j$ matricization of the tensor $\mathcal{R}_i$, and

$$\mathbf{h}_k^{(j)} := \beta_{k,m} \odot \cdots \odot \beta_k^{(j+1)} \odot \beta_k^{(j-1)} \odot \cdots \odot \beta_{k,1} \in \mathbb{R}^{\prod_{j' \neq j} d_{j'}}$$

with $\odot$ the Khatri-Rao product. According to Lemma 3, we can conclude that the minimizer of (A8) is also the solution of

$$\min_{\substack{\beta_{k,j} \\ \|\beta_{k,j}\|_2=1, \|\beta_{k,j}\|_0 \leq s_j}} \left\| [\frac{1}{n} \sum_{i=1}^n \alpha_{ik}^2 \mathcal{R}_i]_{(j)}^\top - \mathbf{h}_k^{(j)} \beta_k^{(j)\top} \right\|_F^2.$$

This is equivalent to solve

$$\min_{\substack{\beta_{k,j} \\ \|\beta_{k,j}\|_2=1, \|\beta_{k,j}\|_0 \leq s_j}} \left\| \frac{1}{n} \sum_{i=1}^n \alpha_{ik}^2 \mathcal{R}_i - w_k \beta_{k,1} \circ \cdots \circ \beta_{k,m} \right\|_F^2.$$

given other parameters $\beta_k^{(j')}$ with $j' \neq j$. Finally, simple algebra leads to the estimator $\widehat{w}_k$. This proves the desirable result in Lemma 1. □

### A3.2 Proof of Lemma 2

Denote $[t_i]_{i_1,\ldots,i_m}$ as the $(i_1,\ldots,i_m)$-th entry of the tensor $\mathcal{T}_i$, and denote $a_{i_1,\ldots,i_m}$ as the $(i_1,\ldots,i_m)$-th entry of the tensor $\mathcal{A}_k$. By the definition of the tensor Frobenius norm, we have

$$\frac{\partial \frac{1}{n} \sum_{i=1}^n \left\| \mathcal{T}_i - \alpha^\top \mathbf{x}_i \mathcal{A}_k \right\|_F^2}{\partial \alpha} = \frac{1}{n} \sum_{i=1}^n \sum_{i_1,\ldots,i_m} 2([t_i]_{i_1,\ldots,i_m} - \alpha^\top \mathbf{x}_i a_{i_1,\ldots,i_m})(-a_{i_1,\ldots,i_m} \mathbf{x}_i).$$

Setting the above gradient to zero implies that

$$\frac{1}{n} \sum_{i=1}^n \Big\{ \sum_{i_1,\ldots,i_m} [t_i]_{i_1,\ldots,i_m} a_{i_1,\ldots,i_m} \Big\} \mathbf{x}_i = \frac{1}{n} \sum_{i=1}^n \sum_{i_1,\ldots,i_m} \alpha^\top \mathbf{x}_i a_{i_1,\ldots,i_m}^2 \mathbf{x}_i,$$

where the left-hand-side equals $n^{-1} \sum_{i=1}^n \|\mathcal{T}_i * \mathcal{A}_k\|_+ \mathbf{x}_i$, and the right-hand-side equals

$$\Big\{ \sum_{i_1,\ldots,i_m} a_{i_1,\ldots,i_m}^2 \Big\} \frac{1}{n} \sum_{i=1}^n \mathbf{x}_i \mathbf{x}_i^\top \alpha = \|\mathcal{A}_k\|_F^2 \frac{1}{n} \sum_{i=1}^n \mathbf{x}_i \mathbf{x}_i^\top \alpha.$$

This proves the desirable conclusion in Lemma 2. □

### A3.3 Proof of Lemma 3

We first derive the solution of

$$\arg\min_{\beta} \frac{1}{n} \sum_{i=1}^n \gamma_i^2 \left\| \mathbf{Y}_i - \alpha \beta^\top \right\|_F^2,$$

and then connect it with the optimization problem on the right-hand-side. Denote $[\mathbf{Y}_i]_{sj} \in \mathbb{R}$ as the $(s,j)$-th entry of the matrix $\mathbf{Y}_i$, and denote $[\mathbf{Y}_i]_j \in \mathbb{R}^{d_1}$ as the $j$-th column of the matrix $\mathbf{Y}_i$. Note that, for each $j = 1, \ldots, d_2$, solving

$$\frac{\partial \frac{1}{n} \sum_{i=1}^n \gamma_i^2 \left\| \mathbf{Y}_i - \alpha \beta^\top \right\|_F^2}{\partial \beta_j} = \frac{1}{n} \sum_{i=1}^n \sum_{s=1}^{d_1} 2\gamma_i^2 ([\mathbf{Y}_i]_{sj} - \alpha_s \beta_j)(-\alpha_s) = 0$$

leads to the solution

$$\widehat{\beta}_j = \frac{\frac{1}{n}\sum_{i=1}^n \sum_{s=1}^{d_1} \gamma_i^2 [\mathbf{Y}_i]_{sj}\alpha_s}{\frac{1}{n}\sum_{i=1}^n \gamma_i^2 \sum_{s=1}^{d_1} \alpha_s^2} = \frac{\frac{1}{n}\sum_{i=1}^n \gamma_i^2 [\mathbf{Y}_i]_j^\top \alpha}{(\frac{1}{n}\sum_{i=1}^n \gamma_i^2)\alpha^\top \alpha}.$$

The solution $\widehat{\beta} = (\widehat{\beta}_1, \ldots, \widehat{\beta}_{d_2})^\top$ is indeed a minimizer to the original optimization problem by noting that the second order derivative of $\beta_j$ is positive. By solving a similar equation, we get the minimizer of the optimization problem $\min_{\beta_j} \left\| \frac{1}{n}\sum_{i=1}^n \mathbf{Y}_i - \alpha\beta^\top \right\|_F^2$ as

$$\widetilde{\beta}_j = \frac{\frac{1}{n}\sum_{i=1}^n \gamma_i^2 [\mathbf{Y}_i]_j^\top \alpha}{\alpha^\top \alpha}.$$

Note that $\widehat{\beta}_j$ equals to $\widetilde{\beta}_j$ up to a constant $n^{-1}\sum_{i=1}^n \gamma_i^2$. Therefore, normalizing $\widehat{\beta}_j$ and $\widetilde{\beta}_j$ to be unit-norm vector leads to the same solution. This completes the proof of Lemma 3. $\square$

### A3.4  Proof of Lemma 7

According to the Cauchy-Schwarz inequality, we have $\|\mathbf{u}\|_1 \leq \sqrt{s}\|\mathbf{u}\|_2 = \sqrt{s}$, and $\|\mathbf{v}\|_1 \leq \sqrt{s}$, $\|\mathbf{w}\|_1 \leq \sqrt{s}$. Therefore, $\|\mathcal{A}\|_1 = \|\mathbf{u} \circ \mathbf{v} \circ \mathbf{w}\|_1 \leq \|\mathbf{u}\|_1 \cdot \|\mathbf{v}\|_1 \cdot \|\mathbf{w}\|_1 \leq s^{3/2}$. $\square$

### A3.5  Proof of Lemma 9

The key step is to verify the conditions in Lemma 8. In our example, the incoherence condition, Assumption 4, trivially holds, since the incoherence parameter $\zeta = 0$. Therefore, the function $f(\epsilon; K, d_0) = C_3 \epsilon^2$ in our example. According to Lemma 8, we have

$$D(\widehat{\boldsymbol{\beta}}_{k,3}, \boldsymbol{\beta}^*_{k,3}) \leq \frac{\sqrt{5} w^*_{\max} C_3 \epsilon^2 + \sqrt{5}\eta(\widehat{\mathcal{E}}, d_0 + s)}{w_j(1 - \epsilon^2) - w^*_{\max} C_3 \epsilon^2 - \eta(\widehat{\mathcal{E}}, d_0 + s)}.$$

We next simplify the denominator $w_j(1-\epsilon^2) - w^*_{\max} C_3 \epsilon^2 - \eta(\widehat{\mathcal{E}}, d_0 + s)$. By the assumption on $\epsilon$, we have

$$w_j(1 - \epsilon^2) - w^*_{\max} C_3 \epsilon^2 \geq w^*_{\min}/2,$$

which, together with the condition on the error tensor $\eta(\widehat{\mathcal{E}}, d_0 + s)$, implies that

$$w_j(1 - \epsilon^2) - w^*_{\max} C_3 \epsilon^2 - \eta(\widehat{\mathcal{E}}, d_0 + s) \geq w^*_{\min}/4.$$

Therefore, we have

$$D(\widehat{\boldsymbol{\beta}}_{k,3}, \boldsymbol{\beta}^*_{k,3}) \leq \frac{4\sqrt{5} w^*_{\max} C_3}{w^*_{\min}} \epsilon^2 + \frac{4\sqrt{5}}{w^*_{\min}}\eta(\widehat{\mathcal{E}}, d_0 + s),$$

which completes the proof of Lemma 9. $\square$

## A3.6 Proof of Lemma 10

According to the explicit form in ($A3$), we have

$$\eta(\widehat{\mathcal{E}}, s) \leq \eta(I_1, s) + \eta(I_2, s).$$

Next we bound the two terms $\eta(I_1, s)$ and $\eta(I_2, s)$ separately.

**Bound** $\eta(I_1, s)$: Denote $\mathbb{S}_q^d = \{\mathbf{v} \in \mathbb{R}^d \mid \|\mathbf{v}\|_2 = 1, \|\mathbf{v}\|_0 \leq q\}$. According to the definiteness of $\eta(\cdot)$, we have

$$\eta(I_1, s) = \sup_{\mathbf{u} \in \mathbb{S}_s^{d_1}, \mathbf{v} \in \mathbb{S}_s^{d_2}, \mathbf{w} \in \mathbb{S}_s^{d_3}} |I_1 \times_1 \mathbf{u} \times_2 \mathbf{v} \times_3 \mathbf{w}|$$

$$\leq \max_i \left| \frac{(\boldsymbol{\beta}_{k,4}^* - \widehat{\boldsymbol{\beta}}_{k,4})^\top \mathbf{x}_i}{\widehat{\boldsymbol{\beta}}_{k,4}^\top \mathbf{x}_i \boldsymbol{\beta}_{k,4}^{*\top} \mathbf{x}_i} \right| \sup_{\mathbf{u} \in \mathbb{S}_s^{d_1}, \mathbf{v} \in \mathbb{S}_s^{d_2}, \mathbf{w} \in \mathbb{S}_s^{d_3}} \left| \left( \frac{1}{n} \sum_{i=1}^n \mathcal{Y}_i \right) \times_1 \mathbf{u} \times_2 \mathbf{v} \times_3 \mathbf{w} \right|.$$

By the Cauchy Schwarz inequality, we have $\|(\boldsymbol{\beta}_{k,4}^* - \widehat{\boldsymbol{\beta}}_{k,4})^\top \mathbf{x}_i\| \leq \|\boldsymbol{\beta}_{k,4}^* - \widehat{\boldsymbol{\beta}}_{k,4}\| \|\mathbf{x}_i\| \leq C_2 \|\boldsymbol{\beta}_{k,4}^* - \widehat{\boldsymbol{\beta}}_{k,4}\| \leq C_2 \epsilon$. Moreover, note that $|\widehat{\boldsymbol{\beta}}_{k,4}^\top \mathbf{x}_i \boldsymbol{\beta}_{k,4}^{*\top} \mathbf{x}_i| = |\boldsymbol{\beta}_{k,4}^{*\top} \mathbf{x}_i + (\widehat{\boldsymbol{\beta}}_{k,4} - \boldsymbol{\beta}_{k,4}^*)^\top \mathbf{x}_i| |\boldsymbol{\beta}_{k,4}^{*\top} \mathbf{x}_i|$. When the element of $\mathbf{x}_i$ is bounded for each $i = 1, \ldots, n$ and the dimension $d_{m+1}$ of $\mathbf{x}_i$ is fixed, we have $\|\mathbf{x}_i\| \leq C_2$. Then by the assumption that $|\boldsymbol{\beta}_{k,4}^{*\top} \mathbf{x}_i| \geq C_3$ and $\|\mathbf{x}_i\| \leq C_2$, we have

$$|\widehat{\boldsymbol{\beta}}_{k,4}^\top \mathbf{x}_i \boldsymbol{\beta}_{k,4}^{*\top} \mathbf{x}_i| \geq (C_3 - C_2 \|\widehat{\boldsymbol{\beta}}_{k,4} - \boldsymbol{\beta}_{k,4}^*\|) C_3 \geq (C_3 - C_2 \epsilon) C_3 \geq C_3^2 / 2,$$

where the last inequality is due to the assumption that $\epsilon \leq C_3 / (2C_2)$. Then,

$$\max_i \left| \frac{(\boldsymbol{\beta}_{k,4}^* - \widehat{\boldsymbol{\beta}}_{k,4})^\top \mathbf{x}_i}{\widehat{\boldsymbol{\beta}}_{k,4}^\top \mathbf{x}_i \boldsymbol{\beta}_{k,4}^{*\top} \mathbf{x}_i} \right| \leq \frac{2 C_2 \epsilon}{C_3^2}.$$

Moreover, according to the true model $\mathcal{Y}_i = w^*(\boldsymbol{\beta}_{k,4}^{*\top} \mathbf{x}_i) \boldsymbol{\beta}_{k,1}^* \circ \boldsymbol{\beta}_{k,2}^* \circ \boldsymbol{\beta}_{k,3}^* + \mathcal{E}_i$,

$$\sup_{\mathbf{u} \in \mathbb{S}_s^{d_1}, \mathbf{v} \in \mathbb{S}_s^{d_2}, \mathbf{w} \in \mathbb{S}_s^{d_3}} \left| \left( \frac{1}{n} \sum_{i=1}^n \mathcal{Y}_i \right) \times_1 \mathbf{u} \times_2 \mathbf{v} \times_3 \mathbf{w} \right|$$

$$\leq \sup_{\mathbf{u} \in \mathbb{S}_s^{d_1}, \mathbf{v} \in \mathbb{S}_s^{d_2}, \mathbf{w} \in \mathbb{S}_s^{d_3}} \left| \left( w^* \left( \boldsymbol{\beta}_{k,4}^{*\top} \frac{1}{n} \sum_{i=1}^n \mathbf{x}_i \right) \boldsymbol{\beta}_{k,1}^* \circ \boldsymbol{\beta}_{k,2}^* \circ \boldsymbol{\beta}_{k,3}^* \right) \times_1 \mathbf{u} \times_2 \mathbf{v} \times_3 \mathbf{w} \right| + \eta(\bar{\mathcal{E}}, s)$$

$$= \eta(\bar{\mathcal{E}}, s),$$

where the last equality is due to the fact that the covariate vector $\mathbf{x}_i$ is centered such that

$\frac{1}{n}\sum_{i=1}^{n} \mathbf{x}_i = \mathbf{0}$. Therefore, we have,

$$\eta(I_1, s) \leq \underbrace{\frac{2C_2 \eta(\bar{\mathcal{E}}, s)}{C_3^2}}_{\kappa_2} \epsilon.$$

**Bound** $\eta(I_2, s)$: By the assumption that, for each $i$, $|\boldsymbol{\beta}_{k,4}^{*\top}\mathbf{x}_i| \geq C_3$, we have that $\max_i(|\boldsymbol{\beta}_{k,4}^{*\top}\mathbf{x}_i|)^{-1} \leq 1/C_3$, and hence

$$\begin{aligned}
\eta(I_2, s) &= \sup_{\mathbf{u}\in\mathbb{S}_s^{d_1}, \mathbf{v}\in\mathbb{S}_s^{d_2}, \mathbf{w}\in\mathbb{S}_s^{d_3}} \left| \left( \frac{1}{n}\sum_{i=1}^{n} \frac{\mathcal{E}_i}{\boldsymbol{\beta}_{k,4}^{*\top}\mathbf{x}_i} \right) \times_1 \mathbf{u} \times_2 \mathbf{v} \times_3 \mathbf{w} \right| \\
&\leq \max_i(|\boldsymbol{\beta}_{k,4}^{*\top}\mathbf{x}_i|)^{-1} \eta\left( \frac{1}{n}\sum_{i=1}^{n} \mathcal{E}_i, s \right) \\
&\leq \frac{1}{C_3} \eta\left( \frac{1}{n}\sum_{i=1}^{n} \mathcal{E}_i, s \right).
\end{aligned}$$

Combining the above two results, we obtain the desirable upper bound of $\eta(\hat{\mathcal{E}}, s)$. □

### A3.7 Proof of Lemma 11

**Bound** $(I)$: For any two tensors $\mathcal{A}, \mathcal{B}$ of the same dimension, the Cauchy-Schwarz inequality implies that $\langle \mathcal{A}, \mathcal{B} \rangle \leq \|\mathcal{A}\|_F \|\mathcal{B}\|_F$. Therefore, we have

$$(I) \leq \frac{\|\widehat{\mathcal{A}}_k - \mathcal{A}_k^*\|_F}{\|\widehat{\mathcal{A}}_k\|_F}.$$

We next simplify the above numerator $\|\widehat{\mathcal{A}}_k - \mathcal{A}_k^*\|_F$. By definition,

$$\begin{aligned}
\|\widehat{\mathcal{A}}_k - \mathcal{A}_k^*\|_F &= \left\| \widehat{w}_k \widehat{\boldsymbol{\beta}}_{k,1} \circ \widehat{\boldsymbol{\beta}}_{k,2} \circ \widehat{\boldsymbol{\beta}}_{k,3} - w_k^* \boldsymbol{\beta}_{k,1}^* \circ \boldsymbol{\beta}_{k,2}^* \circ \boldsymbol{\beta}_{k,3}^* \right\|_F \\
&\leq \underbrace{\left\| \widehat{w}_k \widehat{\boldsymbol{\beta}}_{k,1} \circ \widehat{\boldsymbol{\beta}}_{k,2} \circ \widehat{\boldsymbol{\beta}}_{k,3} - w_k^* \widehat{\boldsymbol{\beta}}_{k,1} \circ \widehat{\boldsymbol{\beta}}_{k,2} \circ \widehat{\boldsymbol{\beta}}_{k,3} \right\|_F}_{I_1} \\
&+ \underbrace{\left\| w_k^* \widehat{\boldsymbol{\beta}}_{k,1} \circ \widehat{\boldsymbol{\beta}}_{k,2} \circ \widehat{\boldsymbol{\beta}}_{k,3} - w_k^* \boldsymbol{\beta}_{k,1}^* \circ \widehat{\boldsymbol{\beta}}_{k,2} \circ \widehat{\boldsymbol{\beta}}_{k,3} \right\|_F}_{I_2} \\
&+ \underbrace{\left\| w_k^* \boldsymbol{\beta}_{k,1}^* \circ \widehat{\boldsymbol{\beta}}_{k,2} \circ \widehat{\boldsymbol{\beta}}_{k,3} - w_k^* \boldsymbol{\beta}_{k,1}^* \circ \boldsymbol{\beta}_{k,2}^* \circ \widehat{\boldsymbol{\beta}}_{k,3} \right\|_F}_{I_3} \\
&+ \underbrace{\left\| w_k^* \boldsymbol{\beta}_{k,1}^* \circ \boldsymbol{\beta}_{k,2}^* \circ \widehat{\boldsymbol{\beta}}_{k,3} - w_k^* \boldsymbol{\beta}_{k,1}^* \circ \boldsymbol{\beta}_{k,2}^* \circ \boldsymbol{\beta}_{k,3}^* \right\|_F}_{I_4}.
\end{aligned}$$

We next bound each term $I_j/\|\widehat{\mathcal{A}}_k\|_F$ for $j = 1, 2, 3, 4$. First, according to the assumptions $|\widehat{w}_k - w_k^*| \leq \epsilon$ and $\epsilon \leq w_{\min}^*/2$, we have $|\widehat{w}_k| \geq w_k^* - \epsilon \geq w_{\min}^*/2$. Therefore, we have

$$\frac{I_1}{\|\widehat{\mathcal{A}}_k\|_F} = \left|\frac{\widehat{w}_k - w_k^*}{\widehat{w}_k}\right| \leq \frac{\epsilon}{|\widehat{w}_k|} \leq \frac{2}{w_{\min}^*}\epsilon.$$

In addition, note that for any vectors $\mathbf{a} \in \mathbb{R}^m, \mathbf{b} \in \mathbb{R}^n$, since the rank of the matrix $\mathbf{a} \circ \mathbf{b}$ is 1, we have $\|\mathbf{a} \circ \mathbf{b}\|_F = \|\mathbf{a} \circ \mathbf{b}\|_2$. This equality and the fact that $\|\boldsymbol{\beta}_{k,j}^*\|_2 = \|\widehat{\boldsymbol{\beta}}_{k,j}\|_2 = 1$ for any $j = 1, 2, 3$ imply that

$$\frac{I_2}{\|\widehat{\mathcal{A}}_k\|_F} = \frac{\left\|w_k^*(\widehat{\boldsymbol{\beta}}_{k,1} - \boldsymbol{\beta}_{k,1}^*) \circ \widehat{\boldsymbol{\beta}}_{k,2} \circ \widehat{\boldsymbol{\beta}}_{k,3}\right\|_F}{\left\|\widehat{w}_k\widehat{\boldsymbol{\beta}}_{k,1} \circ \widehat{\boldsymbol{\beta}}_{k,2} \circ \widehat{\boldsymbol{\beta}}_{k,3}\right\|_F} \leq \frac{|w_k^*|\|\widehat{\boldsymbol{\beta}}_{k,1} - \boldsymbol{\beta}_{k,1}^*\|_2}{|\widehat{w}_k|} \leq 2\sqrt{2}\epsilon,$$

where the last inequality is due to the inequality $|\widehat{w}_k| \geq w_{\min}^*/2$ as we show above, as well as the fact that $D(\mathbf{u}, \mathbf{v}) \leq \min\{\|\mathbf{u} - \mathbf{v}\|_2, \|\mathbf{u} + \mathbf{v}\|_2\} \leq \sqrt{2}D(\mathbf{u}, \mathbf{v})$ for unit vectors $\mathbf{u}, \mathbf{v}$. By applying similar proof techniques, we also have

$$\frac{I_3}{\|\widehat{\mathcal{A}}_k\|_F} \leq 2\sqrt{2}\epsilon; \quad \frac{I_4}{\|\widehat{\mathcal{A}}_k\|_F} \leq 2\sqrt{2}\epsilon.$$

Combing the above four inequalities, we obtain the desirable bound for $(I)$ in $(A6)$.

**Bound** $(II)$: It is easy to see that

$$(II) \leq \frac{\frac{1}{n}\sum_{i=1}^n \langle \mathcal{E}_i, \widehat{\mathcal{A}}_k\rangle \|\Omega \mathbf{x}_i\|_2}{\|\widehat{\mathcal{A}}_k\|_F^2}.$$

According to the assumptions that $\Omega$ is positive definite with bounded eigenvalues and $\|\mathbf{x}_i\| \leq C_2$ for each $i$, we have $\max_i \|\Omega \mathbf{x}_i\|_2$ is bounded by some constant $C_1 > 0$. In addition, according to the Gaussian comparison inequality Lemma 4, as well as the large deviation bound, we have that, with high probability

$$\frac{1}{n}\sum_{i=1}^n \langle \mathcal{E}_i, \widehat{\mathcal{A}}_k\rangle \leq \mathbb{E}\left[\frac{1}{\sqrt{n}}\langle \mathcal{G}, \widehat{\mathcal{A}}_k\rangle\right] \leq \frac{1}{\sqrt{n}}\mathbb{E}\left[\|\mathcal{G}\|_F \|\widehat{\mathcal{A}}_k\|_F\right],$$

for some Gaussian tensor $\mathcal{G} \in \mathbb{R}^{d_1 \times d_2 \times d_3}$ whose entries are i.i.d. standard normal random variables. Therefore, we have, with high probability, the desirable inequality in $(A7)$ holds, where the whole term $\widetilde{C}$ is bounded by noting that $\max_i \|\Omega \mathbf{x}_i\|_2$ is bounded and all entries of tensors $\mathcal{G}$, and $\widehat{\mathcal{A}}_k$ of the same dimension are bounded. $\square$

### A3.8 Proof of Corollary 1

The derivation of $\eta \left( \frac{1}{n} \sum_{i=1}^{n} \mathcal{E}_i, s \right)$ in the Gaussian error tensor scenario consists of three parts. In Stage 1, we apply Lemma 4 to show that

$$\mathbb{P}\left[ \eta \left( \frac{1}{n} \sum_{i=1}^{n} \mathcal{E}_i, s \right) > x \right] \leq \mathbb{P}\left[ n^{-1/2} \eta \left( \mathcal{G}, s \right) > x \right],$$

for some Gaussian tensor $\mathcal{G}$. In Stage 2, we show via the large deviation bound inequality that

$$\mathbb{P}\left( |\eta \left( \mathcal{G}, s \right) - \mathbb{E}[\eta \left( \mathcal{G}, s \right)]| \geq t \right) \leq 2 \exp\left( -\frac{t^2}{2L^2} \right)$$

with some Lipschitz constant $L$. In Stage 3, by incorporating Lemma 6, and exploring the sparsity constraint, we can compute

$$\mathbb{E}[\eta \left( \mathcal{G}, s \right)]| \leq C \sqrt{s^3 \log(d_1 d_2 d_3)},$$

for some constant $C > 0$.

**Stage 1:** According to the definition of $\eta(\cdot)$ in (8), we have

$$\eta \left( \frac{1}{n} \sum_{i=1}^{n} \mathcal{E}_i, s \right) = \sup_{\substack{\|\mathbf{u}\|=\|\mathbf{v}\|=\|\mathbf{w}\|=1 \\ \|\mathbf{u}\|_0 \leq s, \|\mathbf{v}\|_0 \leq s, \|\mathbf{w}\|_0 \leq s}} \left| \frac{1}{n} \sum_{i=1}^{n} \mathcal{E}_i \times_1 \mathbf{u} \times_2 \mathbf{v} \times_3 \mathbf{w} \right|$$

$$= \sup_{\substack{\|\mathbf{u}\|=\|\mathbf{v}\|=\|\mathbf{w}\|=1 \\ \|\mathbf{u}\|_0 \leq s, \|\mathbf{v}\|_0 \leq s, \|\mathbf{w}\|_0 \leq s}} \left\langle \frac{1}{n} \sum_{i=1}^{n} \mathcal{E}_i, \mathbf{u} \circ \mathbf{v} \circ \mathbf{w} \right\rangle.$$

Clearly, for any unit-norm vectors $\mathbf{u}, \mathbf{v}, \mathbf{w}$, when $\mathcal{E}_i, i=1,\ldots,n$ are i.i.d. Gaussian tensors, we have

$$\mathbb{E}\left[ \left\langle \frac{1}{n} \sum_{i=1}^{n} \mathcal{E}_i, \mathbf{u} \circ \mathbf{v} \circ \mathbf{w} \right\rangle \right] = 0.$$

Since the variance of each entry of $\mathcal{E}_i$ is 1, we also have

$$\mathrm{var}\left[ \left\langle \frac{1}{n} \sum_{i=1}^{n} \mathcal{E}_i, \mathbf{u} \circ \mathbf{v} \circ \mathbf{w} \right\rangle \right] = \mathrm{var}\left[ \frac{1}{n} \sum_{i=1}^{n} \langle \mathcal{E}_i, \mathbf{u} \circ \mathbf{v} \circ \mathbf{w} \rangle \right]$$

$$= \frac{1}{n^2} \sum_{i=1}^{n} \sum_{r,s,t} [\mathbf{u} \circ \mathbf{v} \circ \mathbf{w}]_{r,s,t}^2 \mathrm{var}([\mathcal{E}_i]_{r,s,t}) = \frac{\|\mathbf{u} \circ \mathbf{v} \circ \mathbf{w}\|_F^2}{n}.$$

For some Gaussian tensor $\mathcal{G}$ of the same dimension as $\mathcal{E}_i$, it is easy to show that

$$\mathbb{E}\left[\frac{1}{\sqrt{n}}\langle \mathcal{G}, \mathbf{u} \circ \mathbf{v} \circ \mathbf{w}\rangle\right] = 0$$

$$\mathrm{var}\left[\frac{1}{\sqrt{n}}\langle \mathcal{G}, \mathbf{u} \circ \mathbf{v} \circ \mathbf{w}\rangle\right] = \frac{\|\mathbf{u} \circ \mathbf{v} \circ \mathbf{w}\|_F^2}{n}.$$

Moreover, for any $\mathcal{A}_1 := \mathbf{u}_1 \circ \mathbf{v}_1 \circ \mathbf{w}_1$ and $\mathcal{A}_2 := \mathbf{u}_2 \circ \mathbf{v}_2 \circ \mathbf{w}_2$, we have

$$\mathrm{var}\left[\left\langle \frac{1}{n}\sum_{i=1}^{n}\mathcal{E}_i, \mathcal{A}_1 - \mathcal{A}_2\right\rangle\right] = \frac{\|\mathcal{A}_1 - \mathcal{A}_2\|_F}{n} = \mathrm{var}\left[\frac{1}{\sqrt{n}}\langle \mathcal{G}, \mathcal{A}_1 - \mathcal{A}_2\rangle\right].$$

Therefore, according to Lemma 4, we have, for each $x > 0$,

$$\mathbb{P}\left[\eta\left(\frac{1}{n}\sum_{i=1}^{n}\mathcal{E}_i, s\right) > x\right] \leq \mathbb{P}\left[\frac{1}{\sqrt{n}}\eta(\mathcal{G}, s) > x\right]. \tag{A9}$$

**Stage 2:** We show that the function $\eta(\cdot, s)$ is a Lipschitz function in its first argument. For any two tensors $\mathcal{G}_1, \mathcal{G}_2 \in \mathbb{R}^{d_1 \times d_2 \times d_3}$, denote $\mathcal{A}^* = \sup_{\mathcal{A}}\langle \mathcal{G}_1, \mathcal{A}\rangle$. We have

$$\sup_{\mathcal{A}}\langle \mathcal{G}_1, \mathcal{A}\rangle - \sup_{\mathcal{A}}\langle \mathcal{G}_2, \mathcal{A}\rangle \leq \langle \mathcal{G}_1, \mathcal{A}^*\rangle - \sup_{\mathcal{A}}\langle \mathcal{G}_2, \mathcal{A}\rangle \leq \langle \mathcal{G}_1, \mathcal{A}^*\rangle - \langle \mathcal{G}_2, \mathcal{A}^*\rangle \leq \langle \mathcal{G}_1 - \mathcal{G}_2, \mathcal{A}^*\rangle.$$

Therefore, we have

$$|\eta(\mathcal{G}_1, s) - \eta(\mathcal{G}_2, s)|$$
$$= \sup_{\substack{\|\mathbf{u}\|=\|\mathbf{v}\|=\|\mathbf{w}\|=1 \\ \|\mathbf{u}\|_0 \leq s, \|\mathbf{v}\|_0 \leq s, \|\mathbf{w}\|_0 \leq s}} \left|\mathcal{G}_1 \times_1 \mathbf{u} \times_2 \mathbf{v} \times_3 \mathbf{w}\right| - \sup_{\substack{\|\mathbf{u}\|=\|\mathbf{v}\|=\|\mathbf{w}\|=1 \\ \|\mathbf{u}\|_0 \leq s, \|\mathbf{v}\|_0 \leq s, \|\mathbf{w}\|_0 \leq s}} \left|\mathcal{G}_2 \times_1 \mathbf{u} \times_2 \mathbf{v} \times_3 \mathbf{w}\right|$$
$$\leq \sup_{\substack{\|\mathbf{u}\|=\|\mathbf{v}\|=\|\mathbf{w}\|=1 \\ \|\mathbf{u}\|_0 \leq s, \|\mathbf{v}\|_0 \leq s, \|\mathbf{w}\|_0 \leq s}} \left\langle \mathcal{G}_1 - \mathcal{G}_2, \mathbf{u} \circ \mathbf{v} \circ \mathbf{w}\right\rangle$$
$$\leq \sup_{\|\mathbf{u}\|=\|\mathbf{v}\|=\|\mathbf{w}\|=1} \|\mathbf{u} \circ \mathbf{v} \circ \mathbf{w}\|_F \cdot \|\mathcal{G}_1 - \mathcal{G}_2\|_F$$
$$\leq \|\mathcal{G}_1 - \mathcal{G}_2\|_F,$$

where the second inequality is due to the fact that $\langle \mathcal{A}, \mathcal{B}\rangle \leq \|\mathcal{A}\|_F \|\mathcal{B}\|_F$, and the third inequality is because $\|\mathbf{u} \circ \mathbf{v} \circ \mathbf{w}\|_F = \|\mathbf{u} \circ \mathbf{v} \circ \mathbf{w}\|_2 \leq \|\mathbf{u}\|_2 \|\mathbf{v}\|_2 \|\mathbf{w}\|_2 = 1$ for any unit-norm vectors $\mathbf{u}, \mathbf{v}, \mathbf{w}$.

Applying the concentration result of Lipschitz functions of Gaussian random variables in Lemma 5 with $L = 1$, for the Gaussian tensor $\mathcal{G}$, we have

$$\mathbb{P}(|\eta(\mathcal{G}, s) - \mathbb{E}[\eta(\mathcal{G}, s)]| \geq t) \leq 2\exp\left(-\frac{t^2}{2}\right). \tag{A10}$$

**Stage 3:** We aim to bound $\mathbb{E}[\eta(\mathcal{G}, s)]|$. For a tensor $\mathcal{T} \in \mathbb{R}^{d_1 \times d_2 \times d_3}$, denote its $L_1$-norm regularizer $R(\mathcal{T}) = \sum_{j_1} \sum_{j_2} \sum_{j_3} |\mathcal{T}_{j_1, j_2, j_3}|$. Define the ball of this regularizer as $B_R(\delta) := \{\mathcal{T} \in \mathbb{R}^{d_1 \times d_2 \times d_3} | R(\mathcal{T}) \leq \delta\}$.

For any vectors $\mathbf{u} \in \mathbb{R}^{d_1}, \mathbf{v} \in \mathbb{R}^{d_2}, \mathbf{w} \in \mathbb{R}^{d_3}$ satisfying $\|\mathbf{u}\|_2 = \|\mathbf{v}\|_2 = \|\mathbf{w}\|_2 = 1, \|\mathbf{u}\|_0 \leq s, \|\mathbf{v}\|_0 \leq s$, and $\|\mathbf{w}\|_0 \leq s$, denote $\mathcal{A} := \mathbf{u} \circ \mathbf{v} \circ \mathbf{w}$. Lemma 7 implies that $R(\mathcal{A}) \leq s^{3/2}$.

Therefore, we have

$$\mathbb{E}[\eta(\mathcal{G}, s)] = \mathbb{E}\left[\sup_{\substack{\|\mathbf{u}\|=\|\mathbf{v}\|=\|\mathbf{w}\|=1 \\ \|\mathbf{u}\|_0 \leq s, \|\mathbf{v}\|_0 \leq s, \|\mathbf{w}\|_0 \leq s}} \langle \mathcal{G}, \mathbf{u} \circ \mathbf{v} \circ \mathbf{w} \rangle\right]$$

$$\leq \mathbb{E}\left[\sup_{\mathcal{A} \in B_R(s^{3/2})} \langle \mathcal{G}, \mathcal{A} \rangle\right] = s^{3/2} \mathbb{E}\left[\sup_{\mathcal{A} \in B_R(1)} \langle \mathcal{G}, \mathcal{A} \rangle\right].$$

This result, together with Lemma 6, implies that

$$\mathbb{E}[\eta(\mathcal{G}, s)]| \leq C\sqrt{s^3 \log(d_1 d_2 d_3)}. \tag{A11}$$

Finally, combing ($A10$) and ($A11$), and setting $t = \sqrt{s^3 \log(d_1 d_2 d_3)}$, we have, with probability $1 - 2\exp(-s^3 \log(d_1 d_2 d_3)/2)$,

$$|\eta(\mathcal{G}, s) - \mathbb{E}[\eta(\mathcal{G}, s)]| \leq \sqrt{s^3 \log(d_1 d_2 d_3)}.$$

Henceforth

$$\eta(\mathcal{G}, s) \leq (C+1)\sqrt{s^3 \log(d_1 d_2 d_3)},$$

which together with ($A9$), implies that

$$\eta\left(\frac{1}{n} \sum_{i=1}^n \mathcal{E}_i, s\right) = O_p\left(\sqrt{\frac{s^3 \log(d_1 d_2 d_3)}{n}}\right).$$

This completes the proof of Corollary 1. □

### A3.9 Proof of Corollary 2

The proof follows by incorporating the result in Lemma 1 and the structure assumption in (12). In particular, Lemma 1 implies that, when $\widetilde{\mathcal{E}}_i \in \mathbb{R}^{d \times d}$ is a matrix whose entries are i.i.d. standard Gaussian,

$$\eta\left(\frac{1}{n} \sum_{i=1}^n \widetilde{\mathcal{E}}_i, s\right) = O_p\left(\sqrt{\frac{s^2 \log(d^2)}{n}}\right).$$

Moreover, under the structure assumption in (12), we have

$$\eta\left(\frac{1}{n}\sum_{i=1}^{n}\mathcal{E}_i, s\right) = \eta\left(\frac{1}{n}\sum_{i=1}^{n}\widetilde{\mathcal{E}}_i, s\right),$$

which completes the proof of Corollary 2. □